\documentclass[conference,compsoc]{IEEEtran}
\IEEEoverridecommandlockouts

\usepackage{adjustbox}
\usepackage{algorithm}
\usepackage{algorithmicx}
\usepackage{algpseudocode}
\usepackage{amsthm}
\usepackage{amsmath} 
\usepackage{amssymb}
\usepackage{amsfonts}
\usepackage{array}

\usepackage{booktabs}

\usepackage{caption} 
\usepackage{cite}
\usepackage{colortbl}

\usepackage{dsfont}

\usepackage{filecontents}

\usepackage{graphicx} 

\usepackage{listings}
\usepackage{longtable}

\usepackage{multicol}
\usepackage{multirow}

\usepackage{pifont}

\usepackage{setspace}
\usepackage{stfloats}
\usepackage{subcaption}  

\usepackage{textcomp}
\usepackage{tikz} 
\usepackage{titlesec}

\usepackage{ulem}
\usepackage{url} 

\usepackage{xcolor}

\usepackage{epstopdf}

\usepackage{tikz-cd}

\newtheorem{theorem}{Theorem}
\newtheorem{lemma}[theorem]{Lemma}
\newtheorem{corollary}[theorem]{Corollary}
\newtheorem{definition}{Definition}

\newcommand{\ignore}[1]{}

\newcommand\malurl[1]{\href{notalink}{\nolinkurl{#1}}}

\def\BibTeX{{\rm B\kern-.05em{\sc i\kern-.025em b}\kern-.08em
    T\kern-.1667em\lower.7ex\hbox{E}\kern-.125emX}}

\newcommand\green[1]{{\textcolor{green}{#1}}}
\newcommand{\OURMETHOD}{SEAM\ }

\ignore{
s&p page limitation:
less than 13 pages of main 
less than 18 pages of full
}

\date{}

{
\makeatletter

\def\ps@IEEEtitlepagestyle{%
    \def\@oddfoot{\mycopyrightnotice}%
    \def\@evenfoot{}%
}
\def\mycopyrightnotice{%

    {
    \begin{tabular}{@{}l@{}}
    &\footnotesize  © 2023, Rui Zhu. Under license to IEEE
    &\footnotesize  DOI 10.1109/SP46215.2023.00070
    \end{tabular}
    }\\
    \gdef\mycopyrightnotice{}
}
\makeatletter

}

\makeatother

\makeatletter
\newcommand*\titleheader[1]{\gdef\@titleheader{#1}}
\AtBeginDocument{%
  \let\st@red@title\@title%
  \def\@title{%
    \bgroup\normalfont\large\raggedright\@titleheader\par\egroup
    \vskip1.5em\st@red@title}
}
\makeatother

\title{Selective Amnesia: On Efficient, High-Fidelity and Blind Suppression of Backdoor Effects in Trojaned Machine Learning Models
}

\titleheader{\footnotesize 2023 IEEE Symposium on Security and Privacy (SP)}

\begin{document}

\author{
\textup{Rui Zhu$^{\green{\ast}}$\thanks{$^{\green{\ast}}$ The first two authors contributed equally to this work.}, Di Tang$^{\green{\ast}}$, Siyuan Tang, XiaoFeng Wang, Haixu Tang}\\
\normalsize{Indiana University Bloomington}\\
\textup{\normalsize{\{zhu11, tangd, tangsi, xw7, hatang\}@iu.edu}}
}

\maketitle

\begin{abstract}

The extensive applications of deep neural network (DNN) and its increasingly complicated architecture and supply chain make the risk of backdoor attacks more realistic than ever. In such an attack, the adversary either poisons the training data of a DNN model or manipulates its training process to stealthily inject a covert backdoor task, alongside the primary task, so as to strategically misclassify inputs carrying a trigger.
Defending against such an attack, particularly removing the backdoor effect from an infected model, is known to be hard.
For this purpose, prior research either requires a recovered trigger, which is hard to come by, or attempts to fine-tune a model on its primary task, which becomes less effective when the clean data is scarce.
In this paper, we present a simple yet surprisingly effective technique to induce ``selective amnesia'' on a backdoored model. Our approach, called SEAM, has been inspired by the problem of \textit{catastrophic forgetting} (\textit{CF}), a long standing issue in continual learning. Our idea is to retrain a given DNN model on randomly labeled clean data, to induce a CF on the model, leading to a sudden forget on both primary and backdoor tasks; then we recover the primary task by retraining the randomized model on correctly labeled clean data. We analyzed \OURMETHOD by modeling the unlearning process as continual learning and further approximating a DNN using Neural Tangent Kernel for measuring CF. Our analysis shows that our random-labeling approach actually maximizes the CF on an unknown backdoor in the absence of triggered inputs, and also preserves some feature extraction in the network to enable a fast revival of the primary task.  We further evaluated \OURMETHOD on both image processing and Natural Language Processing tasks, under both data contamination and training manipulation attacks, over thousands of models either trained on popular image datasets or provided by the TrojAI competition. Our experiments show that \OURMETHOD vastly outperforms the state-of-the-art unlearning techniques, achieving a high \textit{Fidelity} (measuring the gap between the accuracy of the primary task and that of the backdoor) efficiently (e.g., about 30 times faster than training a model from scratch on the MNIST dataset), with only a small amount of clean data (e.g., with a size of just $0.1\%$ of training data for TrojAI models). 
\vspace{-2mm}
\end{abstract}


\vspace{-2mm}
\section{Introduction}
\vspace{-4mm}
\label{sec:introduction}
With the wide applications of deep neural networks (DNN), e.g., in image classification~\cite{sultana2018advancements}, natural language processing (NLP)~\cite{DBLP:journals/corr/abs-1907-11692}, malware detection~\cite{DBLP:conf/eurosp/XuLDC18}, etc., come increased security risks, due to the complexity of DNN models and their training pipelines, which open new attack surfaces. These models are known to be vulnerable to various attacks, such as adversarial learning~\cite{DBLP:conf/nips/IlyasSTETM19} and backdoor injection~\cite{DBLP:conf/sp/JagielskiOBLNL18}. Particularly, in a backdoor attack, the adversary either contaminates the training data for a model or manipulates its training process~\cite{DBLP:journals/access/GuLDG19}, so as to embed a Trojan backdoor into the model; as a result, the model may apparently fulfill its task as anticipated but actually responds to the presence of a predefined pattern, called a \textit{trigger}, by misclassifying the instance carrying the trigger to a wrong label. For example,  a backdoored face recognition model for biometric authentication could always identify anyone wearing a unique pair of glasses as a privileged user of a critical system~\cite{DBLP:conf/cvpr/WengerPBY0Z21}. 

\vspace{3pt}\noindent\textbf{Challenges in backdoor defense}. Effective control of such a backdoor risk is hard. The most explored avenue is detection, which often relies on recovering triggers from DNN models~\cite{NeuralCleanse}~\cite{zhu2020gangsweep} or identifying their anomaly in the presence of noise~\cite{huster2021top}, inputs carrying triggers~\cite{chen2018detecting}, and others~\cite{liu2019abs}. With some moderate success, backdoor detection is fundamentally challenging: the complexity of the models and the diversity of triggers (size, form, location, similarity to legitimate features, etc.) often render existing techniques (e.g., trigger recovery~\cite{NeuralCleanse}) futile; even a successful detection often comes with a significant performance overhead, making these approaches less scalable~\cite{liu2019abs}.  As a prominent example, a prior study shows that a source-specific backdoor (a trigger only causing misclassification of the images from a certain class) has defeated all existing detection solutions, except the one requiring triggered images (less likely to get in practice)~\cite{DBLP:conf/uss/Tang0TZ21}. 

Following detection is removal of a backdoor from an infected DNN model, which is done through \textit{unlearning}.  Specifically, if the trigger has been recovered, one can retrain the model on correctly labeled inputs carrying the trigger to remove the effect of the backdoor~\cite{NeuralCleanse}. This approach, however, is contingent upon trigger recovery, which is hard in general as mentioned earlier. An alternative is what we call \textit{blind} unlearning, a technique that works on a DNN model regardless of whether it  contains a backdoor, in an attempt to weaken or even eliminate the backdoor effect when it does exist. 
Its efficacy can be measured using \textit{Fidelity} (Section~\ref{Evaluation_section}), a metric we propose to 
determine the gap between the model's accuracy for its 
primary task (\textit{ACC}) and that for its backdoor task (called the \textit{attack success rate} or \textit{ASR}). A high \textit{Fidelity} indicates that the unlearning technique largely preserves the desired classification capability of the model while suppressing its unwanted backdoor effect.   
However, achieving the high \textit{Fidelity} in blind unlearning is nontrivial. All existing approaches rely on fine-tuning of a DNN model over a set of clean data, for the purpose of reinforcing the model's functionality for solving the primary task, thereby implicitly weakening its capability to handle its covert backdoor task  (e.g., Fine-pruning~\cite{FinePruning} and Neural Attention Distillation\ignore{ or NAD}~\cite{NAD}). The problem is that a backdoored model with a well trained primary task often has little room for fine-tuning (small loss with little impact on its weights).  
So the effective of this approach can be limited, as acknowledged by the prior research~\cite{liu2018fine}, particularly when the clean datasets for fine-tuning are small (below 10\%). 

\vspace{3pt}\noindent\textbf{Selective amnesia}. Ideally, blind unlearning should lead to a ``selective amnesia'' for the target model, causing it to \ignore{completely} remove the memory about the hidden backdoor task while keeping that for solving the primary classification task. We believe that this cannot be effectively achieved by the existing approaches, fundamentally due to their lack of means to explicitly forget the unknown backdoor. In the meantime, we found that there exists a surprisingly simple yet effective solution, through inducing a \textit{catastrophic forgetting} (\textit{CF})~\cite{DBLP:conf/aaai/KemkerMAHK18} on a DNN model and then recovering its desired functionality using a task similar to its overt primary task. More specifically, given a model, our approach first retrains it on a set of data with random labels to cause a CF, through which the model forgets both its primary classification task and its hidden backdoor task; then we utilize a small set of clean data to train the model on the primary task, leading to the recovery of the task without revival of the backdoor. This approach, which we call \OURMETHOD (selective amnesia), turns out to be highly effective: on MNIST, GTSRB and CIFAR10 datasets, the backdoored models processed by SEAM achieve a high \textit{Fidelity} when using a clean dataset with a size of 0.1\% of training data for forgetting and 10\% for recovery; on the infected models for the TrojAI competition, a\ignore{n exceedingly small} clean recovery set, just $0.1\%$ of the training data size, is found to be enough to completely suppress those models' backdoor effects, attaining a high \textit{Fidelity}. The experimental findings demonstrate that \OURMETHOD can nearly fully preserve the model's primary functionality and also almost completely remove the backdoor effect.

To understand why this simple approach works so effectively,  we model the backdoor attack as a problem of multi-task learning to analyze the relations between the primary and the covert backdoor tasks and further utilize Neural Tangent Kernel to approximate a backdoored model and measure the CF incurred by a sequence of tasks (\textit{forgetting} and \textit{recovery}). Our analysis shows that our random-labeling task is \textit{optimal} for \textit{forgetting} a hidden backdoor on a given fixed dataset (e.g., a small subset of clean data). 
Further, under the CF induced by the function, we show that 
a recovery process will selectively revive the primary task by training the model on a similar task (even with a much smaller training dataset). 

We further evaluated \OURMETHOD on DNN models with various architectures (ShuffleNet, VGG, ResNet) for different image recognition and NLP tasks on popular datasets (MNIST, CIFAR10, GTSRG, Imagenet, TrojAI datasets, etc.), under different types of backdoor attacks (Reflection~\cite{reflection} and TrojanNet~\cite{trojannet}).  In all these tests, \OURMETHOD achieved a very high \textit{Fidelity}, nearly fully restoring the original model's ACC and completely removing its backdoor effect, often within a few minutes. Also we ran \OURMETHOD against the state-of-the-art unlearning techniques, including Neural Cleanse (NC)~\cite{NeuralCleanse},  Fine-Pruning~\cite{FinePruning} and Neural Attention Distillation (NAD)~\cite{NAD}, and demonstrated that our approach vastly outperforms these solutions: particularly, given only $0.1\%$ of clean training data, \OURMETHOD reported a \textit{Fidelity} around $90\%$ in less than 1 minute, while other approaches took around an hour to get far lower results ($50\%$ to a bit above $70\%$).  
We also analyzed the robustness of our technique against various evasion attempts.  

\vspace{3pt}\noindent\textbf{Contributions}. Our contributions are outlined as follows:

\vspace{2pt}\noindent$\bullet$\textit{ Novel backdoor defense}. We present a new blind unlearning technique that for the first time, utilizes catastrophic forgetting to achieve efficient and \textit{high-\textit{Fidelity}} removal of Trojan backdoors, in the absence of trigger information.   

\vspace{2pt}\noindent$\bullet$\textit{ Theoretical understandings}. We model the backdoor attack as a multi-task learning problem\ignore{, which has never been done before,} and further leverage Neural Tangent Kernel (NTK)\ignore{ overlap matrix} to measure CF and the similarity between the overt primary task and the covert backdoor task. Our analysis proves the optimality of our forgetting task (random labeling) and also helps better understand the limitations of other unlearning techniques.


\vspace{2pt}\noindent$\bullet$\textit{ Extensive experiments}. We report an extensive evaluation of our new technique, on both image recognition and NLP tasks, under various types of backdoor attacks, and also in comparison with state-of-the-art solutions.  Our evaluation shows that the new approach, though simple, is highly effective, vastly  outperforming existing techniques and fully suppressing backdoor effects even in the presence of a small amount of clean data.



\vspace{-2mm}

\ignore{
Our defence methd is based on these following observation:
\begin{itemize}
    \item ASR can reached high enough as the training just begin(shows in a figure) while the ACC are just start to increase.
    \item While the ASR has changed very little since only few epoch from training begins, the Differential analysis shows that in the early stage, trojaned inputs has few neuron has great difference with benign inputs, while other neuron almost stay the same. However, in the late stage of training, it has totally different phenomenon; compared with benign input, inference a trojaned inputs can make almost every neurons has a lot difference with the benign one. It is pretty much like the processure of human brain mermorize things:(********)
    \item few epochs are much more easy to forget.
\end{itemize}

Based on these observation, we start to think, we hypothesis that at the last stage of training(both asr and acc are high), trigger feature is no more an independent feature that are only exist in few neuron to represent trigger feature, instead, trigger features are binded with other useful task features. if we want to remove trigger feature, one way to slove it is to break this bind first. In our proposed method we first breaked this bind, but it also bring some damage to other useful features, so after that, we recovered this damage.

\subsection{forgottten}
Because most trigger is irrelvant to original task,
In order to break this bind, we use a small amount of data, shuffle their label, feed it to our model. The reason why we do this is because a wrong signal from input can push those useful feature to a wrong direction,and this push can easily break this bind.
}

\section{Background}\label{Section_Background}
\vspace{-4mm}
\subsection{Backdoor}
\vspace{-4mm}
A backdoor attack aims to induce strategic misclassification in a DNN model through training data poisoning~\cite{doan2021lira,label_consistent,cheng2020deep} or training process manipulation~\cite{tang2020embarrassingly,garg2020can}.  
The risk of a backdoor attack can be mitigated through detection: discovering triggers using SGD~\cite{NeuralCleanse}, leveraging images carrying triggers~\cite{zhu2020gangsweep} and others~\cite{xiang2020detection}, as mentioned earlier. A prominent example is Neural Cleanse (NC)~\cite{NeuralCleanse}, which searches for the pattern with the anomalously small norm that causes any image with the pattern to be classified into a target label. Once discovered, the pattern can be used to unlearn the model's backdoor, by training it on the correctly labeled inputs carrying the pattern. Unlike NC, which depends on trigger discovery to unlearn a backdoor from a DNN model, Fine-pruning~\cite{FinePruning} and Neural Attention Distillation~\cite{NAD} are designed to directly remove a backdoor from a Trojaned model, without using any information about the trigger. More specifically, Fine-pruning prunes less informative neurons and then finetunes the model~\cite{FinePruning}, in an attempt to directly erase the backdoor effect; NAD first finetunes a given model to create a \textit{teacher} model, which is combined with the original model (the student) through distillation to unlearn a hidden backdoor~\cite{NAD}. 
\vspace{-2mm}
\subsection{Multi-Task Learning and Continual Learning}
\vspace{-4mm}
In multi-task learning (MTL), several learning tasks are solved jointly at the same time, which could outperform the training alone on individual tasks~\cite{wang2020masking,zhou2019legal,zhao2021adjacency}. More specifically, consider several supervised learning tasks in MTL $\tau_t, t\in [T]$ where $\tau_t = \{x_{t_i},y_{t_i}\}^{n_t}_{i=1}$, with $T\in \mathbb{N}^{\star}$ the set of positive integers. Suppose $\mathcal{X}$ is a feature space of interest, $\mathcal{X} \subseteq \mathbb{R}^p$, and $\mathcal{Y}$ is the space of labels, $\mathcal{Y} \subseteq \mathbb{N}$, then $X_{t} \subseteq \mathbb{R}^{n_t \times p}$ ($p$ is the feature dimension) represents the dataset of the task $\tau_t$ and $x_{i}^{t}, i=1, \ldots, n_{t} \in \mathcal{X}$ is a sample with its corresponding label $y_{i}^{t} \in \mathcal{Y}$. The goal of MTL is to learn a function: $f_{\omega}:\mathcal{X} \times \mathcal{T} \rightarrow \mathcal{Y}$ with $\omega \in \mathbb{R}^{q}$ being the parameters  that fit the prediction as accurate as possible. 
As a special case of MTL, Continual Learning (CL)~\cite{xu2018reinforced,nguyen2017variational} solves a stream of supervised learning tasks $\tau_1,\tau_2,...,\tau_T$. The goal is to train a predictor with the best accuracy on each task assuming that the training data for previous tasks are no longer available after the tasks are accomplished.
In our paper, we utilizes the continual learning theory to model the unlearning process used by SEAM. Specifically, we consider the backdoor unlearning as an independent process after the given model (benign or backdoored) has already been trained.
\vspace{-2mm}
\subsection{Neural Tangent Kernel}
\vspace{-4mm}
Neural network is often considered to be a black box model since it may not offer much insight into the function $f(\cdot)$ it approximates. Recently, neural tangent kernel (NTK) has been proposed to provide a set of theoretical tools for understanding neural networks. In the NTK theory, a neural network is approximated by a kernel machine, in which the kernel function is composed of the gradients of neural network parameters with regard to the training data. More specifically, NTK utilizes the Taylor expansion of the network function with respect to the weights around its initialization for the approximation:
\vspace{-5pt}
\begin{equation}\label{NTK_def}
\notag
    f(x, w) \approx f\left(x, w_{0}\right)+\nabla_{w} f\left(x, w_{0}\right)^{T}\left(w-w_{0}\right),
\vspace{-5pt}
\end{equation}
\noindent where $w_{0}$ is the neural network's initial weights, and $\nabla_{w} f\left(x, w_{0}\right)$ is what so called "NTK" and can be represented in the feature map form $\phi(x)=\nabla_{w} f\left(x, w_{0}\right)$
in which $\phi$ is the feature map in the kernel (NTK) space. Note that, once we choose to expand the initial weights $w_0$, the feature map $\phi$ is fixed. This means that the NTK approximation in the following training procedure utilizes the same feature space, but different training data will optimize each feature's weight. 
In practice, it is not necessary to perform the expansion around the initial weights $w_0$. In fact, the Taylor expansion can have better precision if it is performed around the weights in a later state $w_z$ when we approximate the neural network after the state $z$. For more details of the NTK theory, we refer to the prior research~\cite{jacot2018neural}, and a recent implementation of the NTK framework for a convolutional neural network~\cite{arora2019exact}.

\vspace{-3mm}
\subsection{Threat Model}
\vspace{-4mm}

\vspace{3pt}\noindent\textbf{Defender's goal}. Removing the backdoor injected into the target model and keeping the accuracy of the backdoor removed model be close to what of the original model.

\vspace{3pt}\noindent\textbf{Defender's capabilities}. We assume that the defender has access to the backdoored model and a small set of clean data (with no trigger-carrying samples) but does not know information about the trigger. We believe that the availability of the clean data is realistic, given the fact that the user of an ML model today tend to possess a set of clean testing data for evaluating the model's functionality. For example, Paper With Code~\cite{papers_with_code} provides pre-trained ML models, whose qualities can be evaluated by the user on the public benchmarks released by trusted parties~\cite{wu2022backdoorbench}; these benchmarks can serve as the clean dataset. As another example, image classification models tend to be evaluated using ImageNet~\cite{deng2009imagenet}, a public dataset with integrity protection (md5 fingerprints); so even when a model itself is contaminated during its training process, still we can utilize the ImageNet data, which is supposed to be clean, to unlearn its backdoor. We further assume that this small set of the clean data has the same distribution as that of the sample space the primary task is supposed to work in. We make this ``same distribution" assumption since the success of recovery relies on the similarity between the distribution of the samples and that of the clean dataset for recovery. This assumption is also required for the training data on which any machine learning model is being built. 

\vspace{-3pt}
\vspace{3pt}\noindent\textbf{Adversary goal}. The adversary intends to inject a backdoor into the target model through either training data poisoning or training process manipulation.  

\vspace{3pt}\noindent\textbf{Adversarial capabilities}. We assume a white-box adversary who can access both data and model parameters during training, but does not have influence on the way the model is modified by its user after the training is complete and the model is released. Particularly, we assume that the adversary can inject a backdoor into the victim model, but cannot interfere with SEAM's unlearning operations on the model. 
This assumption is reasonable, since when a backdoored model (compromised during its training process) is given to the user, it tends to be outside the adversary's control: otherwise, the adversary can easily tamper with the answer sent back to the user even in the absence of the backdoor, rendering the backdoor attack less meaningful. Also the adversary's capability to pollute the data for the recovery operation or model fine-tuning can be constrained in various real-world scenarios, as the examples described above.

\vspace{-2mm}




\ignore{
Neural network is often considered to be a black box model since it may not offer much insight into the function $F$ it approximates. Recently, neural tangent kernel (NTK) provides a set of theoretical tools for understanding neural networks. In the NTK theory, a neural network is approximated by a kernel machine, in which the kernel function is composed of the gradient of neural network parameters with regard to the training data. 
More specifically, NTK used the Taylor expansion of the network function with respect to the weights around its initialization:
\begin{equation}\label{NTK_def}
    f(x, w) \approx f\left(x, w_{0}\right)+\nabla_{w} f\left(x, w_{0}\right)^{T}\left(w-w_{0}\right)
\end{equation}
\noindent where $w_{0}$ is neural network initial weights, $\nabla_{w} f\left(x, w_{0}\right)$ is what so called "NTK", it can be written as feature map form:
\begin{equation}
    \phi(x)=\nabla_{w} f\left(x, w_{0}\right)
\end{equation}
where $\phi$ represent feature map to the kernel (NTK) space. Note that, once we choose to expand in the initial weights $w_0$, the feature map $\phi$ is fixed, which means, NTK approximation in following training processure, used same feature space, but different training data will bring different optimization of the weight for each feature. In practice, it is not necessary to perform the expansion around the initial weights $w_0$. In fact, the Taylor expansion can have better precision if the it is performed around the weights in a latter state $w_k,k>0$ when we aim to approximate the neural network after the state $k$. For more details of the NTK theory, We refer to the original paper ~\cite{jacot2018neural}, and a recent implementation of the NTK framework for a convolutional neural network (CNN) ~\cite{arora2019exact}
}

\section{Our Method: \OURMETHOD}\label{SEAM_section}
\vspace{-4mm}
\subsection{Motivation}
\vspace{-4mm}



Our research shows that fine-tuning a Trojaned model cannot guarantee removal of its backdoor, particularly when the clean dataset for fine-tuning is small ($<$10\% of the model's original training data). This has been acknowledged in the prior research~\cite{FinePruning},
even with its effort to prune suspicious neurons to improve the efficacy of fine-tuning. 

Fundamentally, we believe that fine-tuning is limited in its potential to suppress the backdoor effect in general: consider a well-trained but backdoored model, with its ACC close to 1; fine-tuning the model on the clean dataset under its primary (overt) task will have little impact on its weights, given that its loss is already small, and therefore will not significantly interfere with its covert backdoor task. 

A natural solution here is to explicitly forget the backdoor information without undermining the model's capability to solve its primary task. This purpose can be served, effectively, by the idea of \textit{Catastrophic Forgetting(CF)}, as discovered in our research.  CF has long been considered to be a problem for artificial neural network (NN)~\cite{kirkpatrick2017overcoming}, causing an NN to completely and abruptly lose memory of the previously learnt task when learning the information about a new task. The problem is known to be a major barrier to continual learning~\cite{hu2018overcoming}, since the weights in an NN for storing the knowledge of one task will be changed to meet the requirements of the subsequently learnt task. This long-standing trouble, however, was leveraged in our research to enhance the trustworthiness of a DNN. More specifically, prior research shows that a CF can be induced by a new task with similar input features as the old task but different classification outputs~\cite{doan2021theoretical}. This property was utilized by us to build a novel pipeline in which a forgetting task is first run to cause the maximum CF (Section~\ref{section_Backdoor Forgetting}), and then a recovery task is performed to revive the primary task without wakening the backdoor. In this way, we can achieve a selective amnesia on an infected DNN to remove its backdoor effect. 


\vspace{-2mm}
\subsection{The \OURMETHOD Pipeline}
\vspace{-4mm}
At the center of our blind unlearning pipeline is the forgetting task for inducing a CF on a given DNN model. The task is meant to achieve the following goals: 1) ensuring a large CF and 2) helping a quick and selective recovery from the CF. For 1), we need to largely preserve input feature extraction but incur significant impacts on classification, so as to interfere with the original tasks, including the backdoor task; for 2), we hope that the changes to the weights of the DNN, as caused by the interference, can be easily and selectively reversed, so the primary task can be recovered. To this end, we designed a \textit{random-labeling} task that assigns a random class label to each output of the model. This simple approach utilizes the features the original model discovers on the input and intermediate layers but causes a large loss that leads to significant changes to the weights of the layers close to the output. Further, such changes can be done with a few rounds of updates on the weights through stochastic gradient descent (SGD), so the impacts on the primary task can be quickly reversed.

\vspace{3pt}\noindent\textbf{Algorithm}. Based upon this simple forgetting task, we developed the \OURMETHOD pipeline, as illustrated in Algorithm~\ref{alg:forbad}. \OURMETHOD has two steps: \textit{forgetting} and \textit{recovery}, and takes the following inputs: an NN $f(\cdot)$, a labeled \textit{forgetting} dataset $\mathcal{D}_{for}$ for the forgetting task, the number of epochs $\mathbf{N}_{for}$ for running the forgetting task, the accuracy threshold $Acc_{for}$ for an early stop at the forgetting step, 
a \textit{recovery} dataset $\mathcal{D}_{rec}$ for recovering the primary task, its training epochs $\mathbf{N}_{rec}$,
and accuracy threshold $Acc_{rec}$ for an early stop. In Algorithm~\ref{alg:forbad}, \textit{Line} 1-9 describe the \textit{forgetting} step: starting from the original model $f(\cdot)$, we randomly re-label each sample in $\mathcal{D}_{for}$ with a class different from its desirable class (randomly wrong class), aiming to build a randomly labeled training dataset $\mathcal{\bar{D}}_{for}$ for each epoch to train the model on $\mathcal{\bar{D}}_{for}$;
after $\mathbf{N}_{for}$ epochs, the resulting model $\bar{f}$ is expected to classify any given input sample to each label with a similar probability. 
\textit{Line} 10-17 show the \textit{recovery} step: we re-train the model $\bar{f}$ on the dataset $\mathcal{D}_{rec}$ for $\mathbf{N}_{rec}$ epochs to revive the primary task. 

\vspace{-0.5em}
\begin{algorithm}[htb]
    \caption{\OURMETHOD}
    \begin{algorithmic}[1]
        \Require{$f(\cdot), \mathcal{D}_{for}, \mathbf{N}_{for}, Acc_{for}, \mathcal{D}_{rec}, \mathbf{N}_{rec}, Acc_{rec}$} 
        \Ensure{$\tilde{f}(\cdot)$}
        \State $\bar{f} \xleftarrow{} f$  \;
        \For{$epoch$ \textbf{in} \text{range($\mathbf{N}_{for}$)}}
        \State  $\bar{\mathcal{D}}_{for} \xleftarrow{} $ Randomly wrong label $\mathcal{D}_{for}$\;
        \State   $\bar{f} \xleftarrow{}$ Train $\bar{f}$ on $\bar{\mathcal{D}}_{for}$\;
        \State $Acc \xleftarrow{} $ Test $\bar{f}$ on $\mathcal{D}_{for}$.
        \If{$Acc <  Acc_{for}$} 
            \State Break\;
        \EndIf 
        \EndFor
        \State $\tilde{f} \xleftarrow{} \bar{f}$  \;
        \For{$epoch$ \textbf{in} \text{range($\mathbf{N}_{rec}$)}}
        \State   $\tilde{f} \xleftarrow{}$ Train $\tilde{f}$ on $\mathcal{D}_{rec}$\;
        \State $Acc \xleftarrow{} $ Test $\tilde{f}$ on $\mathcal{D}_{rec}$.
        \If{$Acc >  Acc_{rec}$ } 
            \State Break\;
        \EndIf 
        \EndFor
        
    \end{algorithmic}
    \label{alg:forbad}
\end{algorithm}
\vspace{-0.5em}

Our research shows that a very small $\mathcal{D}_{for}$, as randomly selected from the clean dataset for $f(\cdot)$, is adequate for almost completely removing backdoor effect from the  model, that is,  resulting in a negligible ASR (Section~\ref{subsec:efficacy}). The threshold $Acc_{for}$ was set in our experiment to $\min(2 / \mathbf{C}, 0.6)$, where $\mathbf{C}$ is the number of classes in the primary task. The \textit{recovery} dataset $\mathcal{D}_{rec}$ includes only correctly labeled data, which could be a subset of the testing data for the primary task. This ensures that the final model $\tilde{f}$ after the \textit{recovery} step achieves an accuracy close to that of the input model $f(\cdot)$, while keeping ASR exceedingly low (approaching 0). Note that there is no overlap between $\mathcal{D}_{rec}$ and $\mathcal{D}_{for}$. In our experiments, we set the threshold $Acc_{rec}$ for the \textit{recovery} step to $0.97$ of the input model's accuracy.



\ignore{

Based upon this simple forgetting task, we developed \OURMETHOD \, as illustrated in Algorithm~\ref{alg:forbad}. \OURMETHOD \ has the two steps: forgetting and recovery, and takes the following inputs: an input neural network $f(\cdot)$, a randomly labeled \textit{forgetting} dataset $\mathcal{D}_{for}$ used to forget the backdoor, 
the number of epochs $\mathbf{N}_{for}$ for forgetting the backdoor,
the accuracy threshold $Acc_{for}$ for early stop in forgetting step, 
a \textit{recovery} dataset $\mathcal{D}_{rec}$ used to recover the primary classification task,  
the number of epochs $\mathbf{N}_{rec}$ for recovering the primary task,
and the accuracy threshold $Acc_{rec}$ for early stop in recovery step. 

In Algorithm~\ref{alg:forbad}, \textit{line} 1-9 describe the \textit{forgetting} step, in which from the initial model $\mathcal{D}_{for}$, we train a model $\bar{\mathcal{D}}_{for}$ using the forgetting dataset $\bar{f}$. After the training of $\mathbf{N}_{for}$ epochs, the resulting model $\bar{f}$ should predict any input sample as each class with about the same probability. \textit{Line} 10-17 describe the \textit{recovery} step, in which we re-train the model $\bar{f}$ using the recovery dataset $\mathcal{D}_{rec}$ for $\mathbf{N}_{rec}$ epochs. 

\begin{algorithm}[htb]
    \caption{\OURMETHOD}
    \begin{algorithmic}[1]
        \Require{$f(\cdot), \mathcal{D}_{for}, \mathbf{N}_{for}, Acc_{for}, \mathcal{D}_{rec}, \mathbf{N}_{rec}, Acc_{rec}$} 
        \Ensure{$\tilde{f}(\cdot)$}
        \State $\bar{f} \xleftarrow{} f$  \;
        \For{$epoch$ \textbf{in} \text{range($\mathbf{N}_{for}$)}}
        \State  $\bar{\mathcal{D}}_{for} \xleftarrow{} $ Randomly label $\mathcal{D}_{for}$\;
        \State   $\bar{f} \xleftarrow{}$ Train $\bar{f}$ on $\bar{\mathcal{D}}_{for}$\;
        \State $Acc \xleftarrow{} $ Test $\bar{f}$ on $\mathcal{D}_{for}$.
        \If{$Acc <  Acc_{for}$} 
            \State Break\;
        \EndIf 
        \EndFor
        \State $\tilde{f} \xleftarrow{} \bar{f}$  \;
        \For{$epoch$ \textbf{in} \text{range($\mathbf{N}_{rec}$)}}
        \State   $\tilde{f} \xleftarrow{}$ Train $\tilde{f}$ on $\mathcal{D}_{rec}$\;
        \State $Acc \xleftarrow{} $ Test $\tilde{f}$ on $\mathcal{D}_{rec}$.
        \If{$Acc >  Acc_{rec}$ } 
            \State Break\;
        \EndIf 
        \EndFor
        
    \end{algorithmic}
    \label{alg:forbad}
\end{algorithm}

Note that a subset of the testing data for the primary task can be used as the \textit{recovery} dataset $\mathcal{D}_{rec}$, in which each is assumed to be correctly labeled according to our threat model. As a result, the final model $\tilde{f}$ after the recovery step is anticipated to achieve high accuracy on the primary task as the input model $f$, while if the input model $f$ contains an backdoor, it is forgotten after our \textit{forgetting} step, i.e., the attack success rate (ASR) of the backdoor becomes almost 0. For the \textit{forgetting} dataset $\mathcal{D}_{for}$, in the Section~\ref{subsec:}, we will demonstrate that a tiny number of samples are enough to make the input model $f(\cdot)$ forget the injected backdoor (i.e., neglectable ASR). For the threshold $Acc_{for}$, in our experiments, we set it as $\min(2 / \mathbf{C}, 0.6)$ where $\mathbf{C}$ is the number of classes in the primary task. For the threshold $Acc_{rec}$, we set it as $0.97$ times of the accuracy of the input model $f(\cdot)$ on the primary task.

\subsection{Motivation}
describe difference between trigger and primary task.
~\cite{SCAn}
We first want to describe of two tasks: trojan task and primary task. Primary task is the ?????/

During research, we found that continue training clean data in a trojan model cannot remove trojan while the continue training data set is small($<10\%$ from the poison model training data set), this obeservation also mention in [fine-purne]. In order to overcome this problem [fine-purne] add a preprocessure before fine-tune, pruning, which delete those "suspecious" neuron. The definition of "suspecious" neuron is those neuron with no any response with fine-tune data. However the relation between neuron responsibility with trojan feature can be complicated. Arbitrarily delete those "suspecious" neuron is not only has no theortical guidence but also depnedent too much on different kind of hyper parameters.

No matter multi-task learning or continual learning, Catastrophic forgetting is always a core difficulty in multitasking learning. However, such difficulties can not only bring challenges to the research of neural network, but also have good aspects. In this paper, we used other aspect of CF  to make it forget the Trojan task and without forgetting the primary task.

\subsection{\OURMETHOD}

Our method \OURMETHOD contains two steps: \textit{forgetting} and \textit{recovery}. As illustrated in Algorithm~\ref{alg:forbad}, \OURMETHOD takes as input five subjects: an input neural network $f(\cdot)$, a randomly labeled {\em forgetting} dataset $\mathcal{D}_{for}$ used to forget the backdoor, a recovery dataset $\mathcal{D}_{rec}$ used to recover the primary classification task, the number of epochs $\mathbf{N}_{for}$ for forgetting the backdoor, and the number of epochs $\mathbf{N}_{rec}$ for recovering the primary task. In Algorithm~\ref{alg:forbad}, \textit{line} 1-5 describe the \textit{forgetting} step, in which from the initial model $\mathcal{D}_{for}$, we train a model $\bar{\mathcal{D}}_{for}$ using the forgetting dataset $\bar{f}$. After the training of $\mathbf{N}_{for}$ epochs, the resulting model $\bar{f}$ should predict any input sample as each class with about the same probability. \textit{Line} 6-9 describe the \textit{recovery} step, in which we re-train the model $\bar{f}$ using the recovery dataset $\mathcal{D}_{rec}$ for $\mathbf{N}_{rec}$ epochs. Note that a subset of the testing data for the primary task can be used as the recovery dataset $\mathcal{D}_{rec}$, in which each is assumed to be correctly labeled according to our threat model. As a result, the final model $\tilde{f}$ after the recovery step is anticipated to achieve high accuracy on the primary task as the input model $f$, while if the input model $f$ contains an backdoor, it is forgotten, i.e., the attack success rate (ASR) of the backdoor becomes almost 0.

\begin{algorithm}[htb]
    \caption{\OURMETHOD}
    \begin{algorithmic}[1]
        \Require{$f(\cdot), \mathcal{D}_{for}, \mathbf{N}_{for}, \mathcal{D}_{rec}, \mathbf{N}_{rec}$ } 
        \Ensure{$\tilde{f}(\cdot)$}
        \State $\bar{f} \xleftarrow{} f$  \;
        \For{$epoch$ \textbf{in} \text{range($\mathbf{N}_{for}$)}}
        \State  $\bar{\mathcal{D}}_{for} \xleftarrow{} $ Randomly label $\mathcal{D}_{for}$\;
        \State   $\bar{f} \xleftarrow{}$ Train $\bar{f}$ on $\bar{\mathcal{D}}_{for}$\;
        \EndFor
        \State $\tilde{f} \xleftarrow{} \bar{f}$  \;
        \For{$epoch$ \textbf{in} \text{range($\mathbf{N}_{rec}$)}}
        \State   $\tilde{f} \xleftarrow{}$ Train $\tilde{f}$ on $\mathcal{D}_{rec}$\;
        \EndFor
        
    \end{algorithmic}
    \label{alg:forbad}
\end{algorithm}

}

\section{Theoretical Analyses}\label{Theoretical Analyses_section}
\vspace{-4mm}
To show why \OURMETHOD works and why it outperforms other state-of-the-art solutions, we present a theoretical analysis. We first model the backdoor attack, i.e., injection of a backdoor into a DNN so that the network will classify any input sample with a trigger to a target class, as 
a multi-task learning (MTL) problem that aims at learning two distinct tasks simultaneously: the overt primary classification task and the covert trigger recognition (backdoor) task. In the SEAM pipeline, the \textit{forgetting} step is to induce Catastrophic Forgetting (CF) on both primary and backdoor tasks by training the backdoored DNN model using randomly labeled datasets, and the \textit{recovery} step is to restore the performance of the primary task (but not the backdoor task) by re-training the model after the CF on a small set of clean data for the primary task. Under the MTL model, we then analyze \OURMETHOD with the Neural Tangent Kernel (NTK) theory that approximates an NN as a kernel function together with a linear classification model in the kernel space~\cite{jacot2018neural}. This separation of the kernel function (for feature extraction) and classification is important for understanding the effectiveness of our random-labeling task, since it is meant to mostly impact the latter while largely preserving the former. 

More specifically, our analysis shows that 1) the \textit{forgetting} step incurs the most effective CF on the trigger recognition task when the trigger is unknown, and 2) the \textit{recovery} step will not revive the ``forgotten'' backdoor task, as guaranteed by the competition between the primary classification task and the trigger recognition task.
It is important to note that although a backdoored model is usually trained for the primary task (overt task) and the backdoor task (covert task) simultaneously, SEAM performs the forgetting step and the recovery step sequentially on a given model that could be a backdoored model (simultaneously trained for the backdoor and the primary tasks) or a benign model (trained only for the primary task). Therefore, our analysis is meant to evaluate the sequential operations of SEAM, regardless of whether the model SEAM works on is trained simultaneously on the backdoor and primary tasks.

\vspace{-4mm}
\subsection{NTK Modeling of Continual Learning}
\vspace{-4mm}
\ignore{As mentioned earlier, }We use the NTK theory for continual learning in our theoretical analyses. \ignore{In a continual learning setup, }We consider a sequence of tasks $\tau_{1}, \tau_{2}, \ldots, \tau_{T}, T \in \mathbb{N}^{*}$, in which each is a supervised learning task with its input and output in the same high dimensional spaces, respectively. Consider the NN trained on the training data labeled for each of the $T$ tasks in the sequential order. 
Prior research~\cite{doan2021theoretical} utilizes NTK to approximate a NN trained for a target task $\tau_T$ with the model trained for a source task $\tau_S$, where tasks $\tau_S$ and $\tau_T$ are any two tasks in a sequence, and $\tau_S$ occurs before $\tau_T$. The NTK approximation from $f_{\tau_{S}}^{\star}(x)$ to $f_{\tau_{T}}^{\star}(x)$ is expressed as: $f_{\tau_{T}}^{\star}(x) \approx f_{\tau_{S}}^{\star}(x)+\left\langle\phi(x), \omega_{\tau_{T}}^{\star}-\omega_{\tau_{S}}^{\star}\right\rangle,$
where $\omega^{\star}_{\tau_S}$ is the final vector of weights after the training on task $\tau_S$, $\left\langle \cdot \right\rangle$ represents the inner product in the kernel space. $\nabla_{\omega} f_{0}(x) = \phi(x)\ \in \mathbb{R}^{q}$ is defined as the NTK feature map for an input sample $x$, where $q$ is the number of weights in the neural networks. Note that in prior research~\cite{doan2021theoretical} and ~\cite{bennani2020generalisation}, the NTK is defined on the Taylor expansion around the initial state of the DNN ($f_0$), but it can be extended to Taylor expansion around any state of the DNN to approximate a subsequent state of DNN in sequential training.

\vspace{-2mm}
\subsection{Measuring Catastrophic Forgetting using NTK}
\vspace{-4mm}
Following, we first present the definition of CF, which is defined over the transition from a source task to a target one, as measured from a given dataset. Then we utilize NTK to describe the CF measurement, so we can separate an NN's feature representation from classification. This is important for analyzing the impacts of \OURMETHOD on CF (e.g., how the forgetting task interferes with classification).

\vspace{3pt}\noindent\textbf{Definition of CF}. The formal definition of the CF from the prior work~\cite{doan2021theoretical}. 

\begin{definition}
Let $\tau_S$ and $\tau_T$ be the source and target tasks, respectively, where $\tau_S$ is trained before $\tau_T$ in a sequence of continual training tasks, and $D_{\tau_S}=(X_{\tau_S}, Y_{\tau_S})$ be the testing set of the source task. Then the CF of the model for the source task $f_{\tau_S}$ after the training of all the subsequent tasks until the target task $\tau_T$ w.r.t. the testing data $D_{\tau_S}$ is defined as:
\vspace{-5pt}
\begin{equation}
\notag
    \begin{aligned}
\Delta^{\tau_{S} \rightarrow \tau_{T}}\left(X_{\tau_{S}}\right) =\sum_{x \in \mathcal{D}_{\tau_{S}}}\left(f_{\tau_{T}}^{\star}(x)-f_{\tau_{S}}^{\star}(x)\right)^{2}
\end{aligned}
\vspace{-5pt}
\end{equation}
\end{definition}
\vspace{-2mm}
\noindent where $f_{\tau_S}^*$ and $f_{\tau_T}^*$ represent the models (after training) for the source and target tasks, respectively. Throughout this paper, CF is always defined on three elements, the two tasks involved in the task transition: the source task and the target task, and a dataset X on which CF is measured.
Note that the CF can be evaluated on any dataset $X$ taken as the input to the source and target models;
even for the same pairs of source and target models, the CF can be different on different input dataset $X$. Therefore, it is defined as a function, i.e.,  $\Delta^{\tau_{S} \rightarrow \tau_{T}}\left(\cdot \right)$. 
 In the case of multi-class classification, we represent the predicted output as a one-hot vector in which $y_{k}=1$ for the predicted class $k$ and $y_{i}=0$ for all other classes. 
To measure the CF on the transition of such tasks over a dataset $X$, we compute the squared
norm of the one-hot vectors produced by the models for the source and the target tasks, respectively.

\vspace{3pt}\noindent\textbf{Impact on classification: residual}. Interestingly, the above expression of CF is general and has a linear form under the NTK representation, which enables us to perform insightful analyses on the effectiveness of different forgetting approaches. For this purpose, we introduce Lemma 4.1, which is used by the prior study to measure CF~\cite{bennani2020generalisation}:
\begin{lemma}
\label{lemma_cf}
Let $\left\{\omega_{\tau}^{\star}, \forall \tau \in[T]\right\}$ be the weight at the end of the training of task $\tau$. The CF of a source task $\tau_S$ with respect to a target task $\tau_T$ is measured on a data $X$ as:
\vspace{-5pt}
\begin{equation} \label{lemma CF}
    \begin{aligned}
\Delta^{\tau_{S} \rightarrow \tau_{T}}\left(X\right) =\left\|\phi\left(X\right)\left(\omega_{\tau_{T}}^{*}-\omega_{\tau_{S}}^{*}\right)\right\|_{2}^{2}
\end{aligned}
\vspace{-5pt}
\end{equation}
\end{lemma}
Further, as shown in the prior work~\cite{bennani2020generalisation}, we have
\vspace{-5pt}
\begin{equation}\label{Bennani}
    \omega_{\tau_{T}}^{\star}-\omega_{\tau_{S}}^{\star}=\phi\left(X_{\tau_T}\right)^{\top}\left[\phi\left(X_{\tau_T}\right) \phi\left(X_{\tau_T}\right)^{\top}+\lambda I\right]^{-1} \tilde{y}_{\tau_T}
\vspace{-4pt}
\end{equation}
\noindent where $\tilde{y}_{\tau_T} = {y}_{\tau_T} -f^{\star}_{\tau_S}(X_{\tau_T})$ is the {\em residual} between the true labels (i.e., the desirable outputs) and the predicted labels by the source model on the target task's training data $X^{\tau_T}$, which is a vector with the size of the number of samples in $X^{\tau_T}$. $\lambda I$ is the regularization term for better lazy training to improve the precision of the Taylor expansion~\cite{chizat2018lazy}. Note that the residual here describes the impact of the tasks on the classification component of the DNN model while the remaining of Eq.~\ref{Bennani} shows the changes to other components.

\vspace{3pt}\noindent\textbf{Impact on representation: task similarity}. The combination of Eq.~\ref{lemma CF} and Eq.~\ref{Bennani} leads to the following corollary, which is also provided by the prior study~\cite{doan2021theoretical}. It measures the CF incurred by the transition from the task $\tau_S$ to the task $\tau_T$ w.r.t.  $D_{\tau_S}$:
\begin{corollary} \label{coro:similarity}
The CF caused by the sequence of the tasks that end with $\tau_T$ to $\tau_S$ w.r.t. $D_{\tau_S}$ can be expressed as follows: 
\vspace{-5pt}
\begin{equation}
\label{similarity}
\hspace{-2mm}
\scriptscriptstyle
\Delta^{\tau_S \rightarrow \tau_T}\left(X_{\tau_S}\right) 
= \left\| U_{{\tau_S}} \Sigma_{{\tau_S}} \underbrace{V_{{\tau_S}}^{\top} V_{\tau_T}}_{O_{\tau_{S,T}}} \Sigma_{\tau_T}\left[\Sigma_{\tau_T}^{2}+\lambda I\right]^{-1} U_{\tau_T}^{\top} \tilde{y}_{\tau_T}\right\|_{2}^{2}
\end{equation}
\vspace{-8pt}
\end{corollary}
Corollary~\ref{coro:similarity} is extended from the Lemma 1 in paper~\cite{doan2021theoretical}.
This new form of the CF measurement lays the foundation for the analysis on both the forgetting and the recovery tasks. $U,\sigma$ and $V$ represent the left singular vector, singular value, and right singular vector respectively after SVD. Their subscripts represent the tasks they are corresponding to. In addition to the residual for the influence on classification, the term $||O_{\tau_{S,T}}||^2_2=||V_{\tau_{S}}^{\top} V_{\tau_T}||^2_2$, which is positively related to the CF, is considered to be a good {\em similarity} metric between $\tau_S$ and $\tau_T$ in prior studies~\cite{doan2021theoretical}: intuitively, a large $||O_{\tau_{S,T}}||^2_2$ indicates that the angle between the representation vectors produced by the models (for $\tau_S$ and $\tau_T$) on the same input samples (i.e., $X_{\tau_S}$) is small (i.e., these two tasks are similar). Given a fixed residual, a large similarity leads to a large CF, i.e., a high impact of $\tau_T$ on $\tau_{S}$. In Eq.~\ref{similarity}, $U_{\tau_S}$, $\Sigma_{\tau_S}$ and $V_{\tau_S}^{\top}$ result from the SVD of kernel matrix over the dataset $X_{\tau_S}$; the subscript of $\tau_S$ is used here to denote the dataset $X_{\tau_S}$ instead of the source task $\tau_S$. Notably, once the model representation (i.e., the feature map) is fixed for a neural network, the term $O_{\tau_{S, T}}$ is dependent on the dataset $X_{\tau_S}$ (on which the CF is evaluated) as well as the training dataset $X_{\tau_T}$ for the target task. For the rest of the paper, we will refer to this term as the "task similarity" following the terminology in the previous studies~\cite{doan2021theoretical} even though it is not directly related to the two tasks involved in the transition, but to their training datasets. 

\vspace{-2mm}
\subsection{Analyzing Backdoor Forgetting with \OURMETHOD}\label{section_Backdoor Forgetting}
\vspace{-3mm}
Using NTK to model CF, we are able to prove that the random-labeling step of \OURMETHOD maximizes the residual, which is proportional to the CF, on a given training dataset. This result demonstrates that our forgetting task is the best we could do to disrupt the backdoor task in the absence of the information about the backdoor (trigger, source and target classes, etc.).  Following, we elaborate the analysis.

\vspace{3pt}\noindent\textbf{Effectiveness of random labeling}. Suppose that a source NN $f_{\tau_P}^{*}$ is trained on the task $\tau_P$ using a poisoned dataset $D_p = \{X_p,Y_p\}$. The model may perform both the primary classification task and the covert trigger recognition task, because $D_p$ contains a subset of training samples carrying a trigger, denoted as $D_t = \{X_t,Y_t\}$ where every $y_t \in Y_t$ represents the same class (i.e., the target class). 
The goal of the backdoor forgetting is to train a \textit{competitive} task $\tau_F$ starting from the source model $f_{\tau_P}^{*}$ using a dataset $D_{\tau_F}$ such that the backdoor (i.e., the trigger recognition task) injected into the source is forgotten (i.e., unlearned) in the resulting model $f_{\tau_F}^{*}$. According to the NTK analyses described above (Corollary ~\ref{coro:similarity}), for any input data $X$ ($X$ may be a normal input or a trigger-containing input), the CF from the source model for the task $\tau_P$ to the competitive model for the task $\tau_F$ measured on $X$ is:
\begin{equation} \label{P-F forget}
\scriptscriptstyle
\Delta^{\tau_P \rightarrow \tau_F}(X )\\
=\left\| \phi(X) \phi\left(X_{\tau_F}\right)^{\top}\left[\phi\left(X_{\tau_F}\right)  \phi\left(X_{\tau_F}\right)^{\top}+\lambda I\right]^{-1} \tilde{y}_{\tau_F} \right\|_{2}^{2}
\end{equation}
\noindent where $D_{\tau_F} = (X_{\tau_F}, Y_{\tau_F})$ are the labeled training dataset for the task $\tau_F$. Here, the kernel map $\phi(\cdot)$ is defined through the Taylor expansion around the weights of the source model $\omega^{\star}_S$, and remain the same throughout our theoretical analyses of the SEAM approach. Note that, on the right-hand side of Eq.~\ref{P-F forget}, every term after $\phi(X)$ is independent of $X$, whereas only the residual term is dependent on the label (i.e., the desirable output) $y_{\tau_F}$ of the training data $D_{\tau_F}$, and the other terms are dependent only on the input (i.e., $y_{\tau_F}$) of the training data. As a result, we have the following lemma:
\begin{lemma} \label{forget lemma}
For any specific sample $X$, the CF from the source model $f^{\star}_{\tau_P}$ to a competitive model trained on $X_{\tau_F}$ is proportional to the {\em residual}: $\tilde{y}_{\tau_F} = {y}_{\tau_F} -f^{\star}_{\tau_P}(X_{\tau_F})$.
\end{lemma}
Note that this residual is independent of the sample $X$ on which the CF is evaluated. Therefore, we have the following theorem:
\begin{theorem} \label{forget thm}
Given a fixed input of a training dataset $X_{\tau_F}$, the randomly assigned wrong label ${y}_{\tau_F}$ maximizes the residual $\tilde{y}_{\tau_F}$, and thus maximizes the CF of any input sample $X$ from the source model to the competitive model trained on the labeled dataset $D_{\tau_F} = (X_{\tau_F}, y_{\tau_F})$.
\end{theorem}


\vspace{3pt}\noindent\textbf{Discussion}. The proof of Theorem \ref{forget thm} is given in the Appendix. The theorem indicates that if we want to leverage a given dataset $X_{\tau_F}$ to train a \textit{competitive} task $\tau_F$ that maximizes the CF of the task $\tau_P$ on any specific sample $X$, we should resort to the task that maximizes $\tilde{y}_{\tau_F}$, which can be achieved by using a random label ${y}_{\tau_F}$ that is different from their predicted label by the source model $f^{\star}_{\tau_P}(X^{\tau_S})$.
In practice, using a small subset of clean data $D_c = \{X_c,Y_c\}$, we may construct the training dataset $D_{\tau_F}=\{X_c,Y_F\}$ by generating a randomly wrong label for each sample $x \in X_c$ to replace $Y_c$.
Notably, according to Theorem \ref{forget thm}, given a fixed training dataset for the forgetting task $D_{\tau_F}$ (which contains only the clean data without triggers), the model trained on the dataset \ignore{without} with random wrong labels will still induce the maximum CF on any sample $X$ by the source model $f_{\tau_F}^{*}$, no matter whether $X$ is a clean sample set or a trigger-containing sample set. As such, any backdoor task as well as the primary task will be forgotten by our approach. This is the best we can achieve when we do not know the trigger and the backdoor task. On the other hand, if the trigger is known (which is not realistic in a practical scenario), we may selectively unlearn the backdoor task. 


Eq.~\ref{P-F forget} also implies that the direct fine-tuning of the model $f_{\tau_P}^{*}$ using the clean dataset $D_C$ (equivalently, using the primary task as the \textit{competitive} task) will not cause effective unlearning of the backdoor, which is consistent with the findings reported by previous studies~\cite{FinePruning}.
Specifically, during the fine-tuning, the true labels $Y_c$ of the clean samples in $D_c = \{X_c,Y_c\}$ are used, and thus 
the residual $\tilde{y}_{\tau_F} = {Y}_{\tau_F} -f^{\star}_{\tau_P}(X_{\tau_F})$ is very small, because the model $f^{\star}_{\tau_P}$ is expected to correctly predict most of the true labels of the clean data for the primary task. Therefore, the CF induced by fine-tuning is significantly lower than the CF incurred by the \textit{competitive} task of \OURMETHOD trained on randomly labeled training data. 
Intuitively, since the initial model $f_{\tau_P}^{*}$ already makes a correct prediction on most samples in the clean dataset, fine-tuning does not change the model significantly and thus the backdoor task may not be effectively unlearnt. 


%

\begin{figure}[h]
\centering
\includegraphics[width=0.28\textwidth]{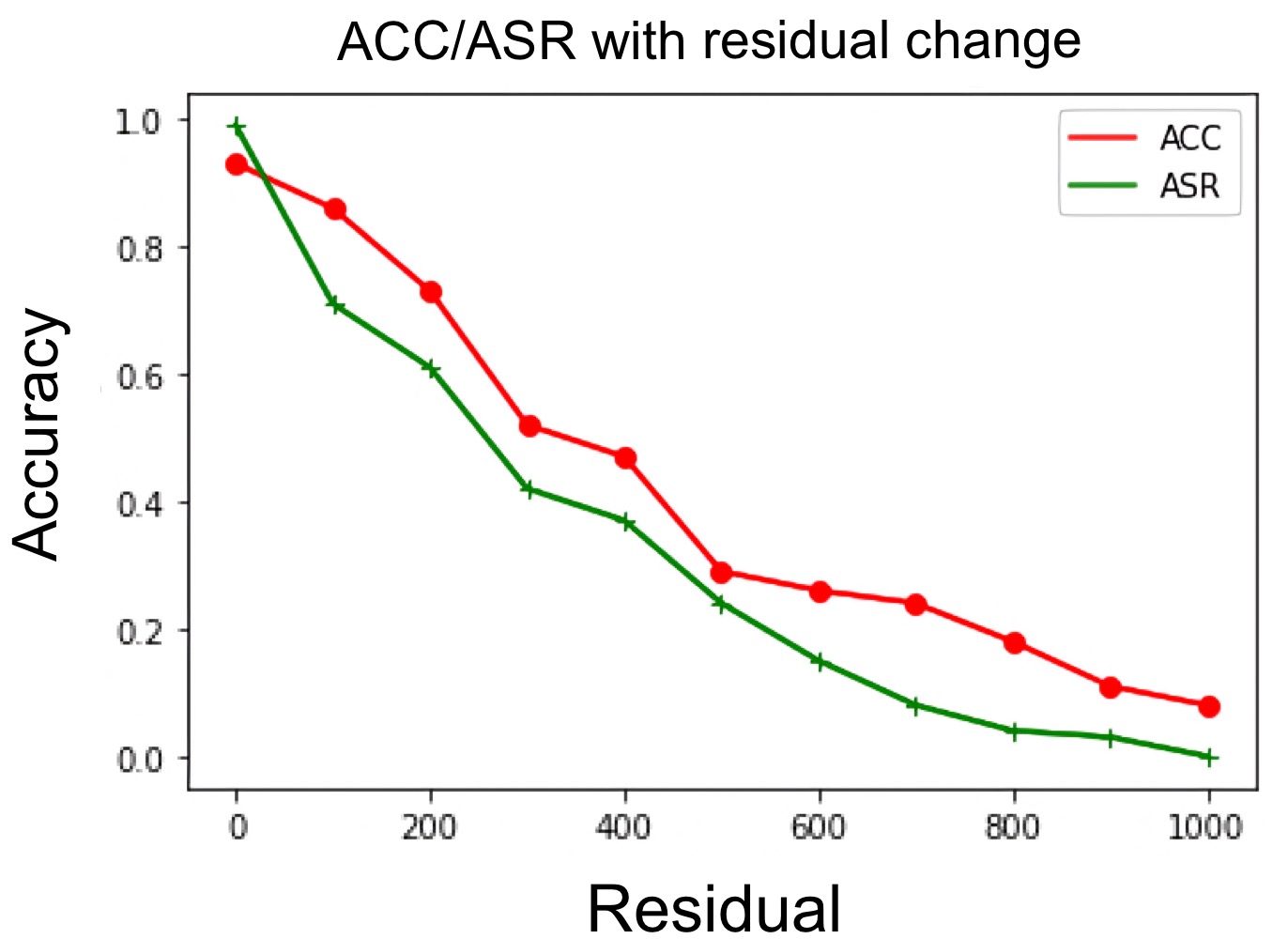}
\caption{Relation of accuracy and residual: The decreasing ACC and ASR with the increasing residual term.}
\label{fig:ACC_ASR_residual}
\vspace{-5pt}
\end{figure}

\vspace{3pt}\noindent\textbf{Example}. 
We conducted a simple experiment on CIFAR-10 to validate our theoretical analyses. Using a training dataset with 50,000 samples, including 2500 trigger-containing ones, we trained a VGG-16 NN with an ACC of $92\%$ for the primary classification task and an ASR of $99\%$ for a polygon trigger. We then ran the \textit{forgetting} step of \OURMETHOD 10 different times on the NN by retraining the model on 1000 clean samples for \textit{only one epoch} (batch size = 256), in which for each time, a different fraction (i.e., 100, 200, ..., 1000, respectively) of the 1000 clean inputs were assigned to randomly wrong labels; as a result, the residual term increased in these 10 experiments. 
The change of ACC and ASR in these 10 forgetting experiments are illustrated in Fig.~\ref{fig:ACC_ASR_residual}. As expected from our theoretical analyses (Theorem~\ref{forget thm}), with the residual term continuing to increase, the ACC and the ASR both decrease to almost zero (i.e., the CF of both the primary and the backdoor tasks becomes maximum). Interestingly, the ASR decreases slightly faster than the ACC, suggesting that in practice, unlearning of the backdoor task might be done more effectively than unlearning of the primary task. 

\vspace{-5pt}

\vspace{-3mm}
\subsection{Analyzing Primary Task Recovery with \OURMETHOD}
\vspace{-4mm}
\label{subsec:analysis_recovery}

Our analysis using the NTK model of the forgetting task shows that \OURMETHOD can maximize the disruption (CF) of an unknown backdoor task for a given set of clean data for training the forgetting task. Following we further report our analysis on the recovery task using NTK. 


\vspace{3pt}\noindent\textbf{Selective recovery}. We denote the recovery task as $\tau_C$, which is trained on a small subset of clean data $D_C=(X_C, Y_C)$ for the primary classification task. According to Corollary~\ref{coro:similarity}, the CF of the \textit{competitive task} $\tau_F$ on any dataset $X$ after learning the recovery task $\tau_C$ is:
\begin{equation} \label{F-R forget}
\scriptscriptstyle
\Delta^{\tau_F \rightarrow \tau_C}\left(X\right) \\
=\left \| U_X \Sigma_X V_X^{\top} V_{\tau_C} \Sigma_{\tau_C}\left[\Sigma_{\tau_C}^{2}+\lambda I\right]^{-1} U_{\tau_C}^{\top} \tilde{y}_{\tau_C}\right\|_{2}^{2}~ 
\end{equation}
\noindent where $U_X \Sigma_X V_X^{\top}$ represents the SVD of the kernel matrix of $X$. Note that the residual $\tilde{y}_{\tau_C}$ represents the difference between the given labels of the recovery data ${X_C}$ and their labels predicted by the \textit{competitive} task. Since the \textit{competitive} task $\tau_F$ is trained on a randomly wrong label dataset $D_{\tau_F}$, it always outputs a randomly wrong label for any given input sample.  Also note that the norm of the residual, $|| \tilde{y}_{\tau_C}||^2$, is expected to be independent (see details in Footnote 2, our extended proof online) of the target task $\tau_C$, as long as the labels of $X_{\tau_C}$ are consistent with $f_{\tau_{P}}^{\star}(X_{\tau_C})$, no matter whether it is the primary task or a backdoor task. Therefore, the following theorem can be proved following the previous work \cite{doan2021theoretical}.


\begin{theorem}\label{recover thm}
If $\tau_F$ is trained on a randomly labeled dataset, 
%
the CF of the task $\tau_F$ on any dataset $X$ after training on the recovery data $D_C$ is dependent on $||O_{X,C}||^2_2=\|V_{X}^{\top} V_{\tau_C}\|^2_2$, i.e., the similarity between representation of the dataset $X$ and the representation of the recovery dataset $X_{\tau_C}$.
\end{theorem}

According to the Theorem~\ref{recover thm}, if $X$ is a subset of the testing dataset for the primary task (i.e., following the same distribution as $D_C$), the representations of the samples in $X$ and $D_C$ are highly similar, and thus $||O_{X,C}||^2_2$ is large, then 
the CF is large, which indicates the recovered model gives very different output from the initial model for the task $\tau_F$ (i.e., it effectively recovers the primary task). In contrast, if $X$ is a dataset for the backdoor task, $||O_{X,C}||^2_2$ is small and thus the CF is small, which indicates the recovered model does not recover the backdoor task.
\vspace{-4mm}

\section{Evaluation}\label{Evaluation_section}
\vspace{-4mm}
We evaluated the effectiveness of our \OURMETHOD pipeline on four datasets (Appendix section 2) against three representative backdoor attacks (Section~\ref{subsec:efficacy}), and also tested its efficiency in terms of its scalability and efficacy using only a very small set of clean data. We further compared our approach with three state-of-the-art unlearning approaches (Section~\ref{subsec:comparison}) and demonstrated the robustness of \OURMETHOD against adaptive evasion (Section~\ref{subsec:evasion}). This evaluation study is based upon the following metrics:

\vspace{3pt}\noindent\textit{ACC for accuracy}. ACC measures the ratio of clean inputs that can be correctly classified. In our study, ACCs are measured on the testing data of each dataset we used. 

\vspace{3pt}\noindent\textit{ASR for attack success rate}. ASR measures the ratio of the inputs carrying triggers that can be misclassified into the target class, when they come from certain source class(es). In our experiments, ASRs are measured on all the images from the source classes in the testing set of each dataset.

\vspace{3pt}\noindent\textit{FID for \textit{Fidelity}}. FID measures the gap between ACC and ASR achieved by a backdoor unlearning technique w.r.t. the original ACC of a backdoored model: $(ACC_s-ASR_s)/ACC_b$ where $ACC_b$ is the model's ACC before the unlearning, and $ACC_s$ and $ASR_s$ are the ACC and ASR after the unlearning. Essentially, FID is the normalized gap between ACC and ASR. The larger such a gap is, the more effective the unlearning process becomes in suppressing ASR and preserving ACC.
We used four public datasets in our experiments: MNIST~\cite{lecun1998gradient}, CIFAR10~\cite{krizhevsky2009learning}, GTSRB~\cite{Stallkamp2012} and the dataset of NIST's TrojAI Competition~\cite{iarpa}. The details of these datasets we used are in Appendix section 2.

\vspace{-2mm}
\subsection{Efficacy}
\vspace{-4mm}
\label{subsec:efficacy}

Here we present our experimental results on the efficacy of \OURMETHOD using the aforementioned datasets. Our approach is found to be highly effective in preserving the ACC of the original model and in the meantime suppressing the effects of backdoors when they are present in the model. In our experiments, we ensure that the testing set used to measure the performance of SEAM does not have any overlap with the clean dataset for inducing CF to the model and recovering its preliminary task, and also the clean set is selected independently from the training dataset (except our study on natural backdoor). In practice, however, the defender could utilize her testing dataset for evaluating the functionality of the model to blindly unlearn the backdoor from the model through SEAM.

\vspace{3pt}\noindent\textbf{On image tasks}. We evaluated \OURMETHOD on the image tasks using MNIST, CIFAR10, GTSRB and the datasets of \textit{Round 3-4} of the TrojAI competition. For each of the MNIST, CIFAR10 and GTSRB datasets, we trained two sets of four DNN models, including ShuffleNet, VGG16, ResNet18 and ResNet101, and ran each set under one of two representative backdoor attacks: Reflection~\cite{reflection}, which is typical of backdoor injection through data poisoning, and TrojanNet~\cite{trojannet}, which is typical of model infection attacks~\cite{backdoor_overview}. On \textit{Round 3-4} datasets of the TrojAI competition, backdoors with various polygon or filter triggers were injected into the models through data poisoning, in which a certain portion of trigger-carrying images were added to the training dataset of each victim model~\cite{iarpa}. 

For the two representative backdoor attacks, Reflection involves selecting a set of candidate images and utilizing reflection transformation to generate triggers, and further injecting them into the training dataset of the victim model. In our experiments, following the steps provided by the Reflection paper~\cite{reflection}, we sampled 200 images from the target class as candidate images and selected the reflection-transformed images with the highest ASRs on a dummy dataset (the training set in our experiments to give the adversary advantage) as triggers, and further pasted such triggers onto 10\% of the selected images and injected them into the training dataset for each victim model. The TrojanNet attack trains a simple NN to recognize whether a trigger appears on the input images and then merges the NN and the victim model~\cite{trojannet}. Again, following the experimental setting given by the prior research~\cite{trojannet}, we utilized a 4x4 square trigger with 11 white pixels and 5 black pixels, and a 4-layer perceptron NN as the trigger recognition network, and further integrated the network into the victim DNN by combining the outputs of their penultimate layers through interpolation. 
Note that both Relection~\cite{reflection} and TrojanNet~\cite{trojannet} requires a source class label as their inputs for generating Trojaned models. \ignore{In the experiments, }We randomly selected a label as the source for each attack.

\ignore{
epoch: depends on dataset (mnist 250, gtsrb 600, cifar 100)
batch size: resnet101 16, other 96

Rf parameters:
learning rate: 0.01
poison rate: 0.1
candidate images: 50 sampled train images
select images: 10 from candidate images
inner epoch: 5 (epoch to choose the best trigger)

Tj parameters:
learning rate: 0.01
marker: 3x3 square white
select\_point: 2
mlp\_dim: 37

CL parameters:
learning rate: 0.01
poison rate: 0.25
poison method: pgd
noise dimension: 100
}

In our experiments on \OURMETHOD against Reflection and TrojanNet over MNIST, GTSRB and CIFAR10, we utilized 0.1\% of the training data of each dataset for the \textit{forgetting} step (label randomization) and 10\% of its training data for the \textit{recovery} step (retraining the randomized model on the primary task). Further, we leveraged the testing data provided by these datasets to measure ACC, and added triggers to all the testing inputs from the source class of each attack to measure ASR. Table~\ref{tab:evaluation_results} presents our experimental results. Here, A high ACC and a high ASR before \OURMETHOD characterize an effective backdoor attack. A high ACC and a low ASR after the unlearning, that is, a high FID, indicate effective suppression of the backdoor effect from a victim model. On all three datasets, \OURMETHOD is found to achieve a good FID against both attacks. The minimum FID is $95.24\%$ for the Trojaned Vgg16 on CIFAR10 under TrojanNet, which is caused by the remaining ASR after unlearning ($3.67\%$). Note that the ACC of Vgg16 on CIFAR10 is among the lowest before the unlearning operation, with and without the backdoor, which indicates the limitation of the model itself and could have an impact on the effectiveness of our approach on the model.


The datasets of TrojAI \textit{Round 3-4} contain a large number of models (3198 models) in various model architectures and with different triggers (Section~\ref{subsec:datasets}). In our experiment, we utilized a very small set of clean data provided by the competition organizer for the \textit{forgetting} and the \textit{recovery} steps, which is only $0.1\%$ of the data for training each model; we also followed the competition's documents~\cite{iarpa} to generate the testing data (a separate dataset about $10\%$ the size of the training data, with those with the source label also used with the trigger for measuring ASR). Table~\ref{tab:evaluation_trojai} illustrates the experiment results. With this unprecedentedly small set of clean data, still \OURMETHOD achieved a FID of $89.85\%$ and $86.75\%$ for \textit{Round 3} and \textit{4}. Actually, only 3-5 clean samples from each class are provided for each model (Section~\ref{subsec:efficiency}). No existing technique, up to our knowledge, could utilize such a small set of clean data for meaningful unlearning (see Table~\ref{tab:evaluation_defenses}).




\begin{table*}[htp]
\centering
\caption{Effectiveness of \OURMETHOD against backdoor attacks on MNIST, GTSRB and CIFAR10. Rf represents Reflection attack, Tj represents TrojanNet attack, $ACC_b$ and $ASR_b$ represents the ACC and ASR after attacks and before \OURMETHOD, $ACC_s$ and $ASR_s$ represents the ACC and ASR after \OURMETHOD, $FID$ represents the \textit{Fidelity} $\frac{ACC_s-ASR_s}{ACC_b}$. \OURMETHOD forgets on 0.1\% training data and recovers on 10\% training data.}
\label{tab:evaluation_results}
\begin{adjustbox}{width=0.85\textwidth}
\begin{tabular}{|c|c|c|cc|cc|cc|cc|cc|}
\hline
\multirow{2}{*}{Dataset} & \multirow{2}{*}{Model} & \multirow{2}{*}{ACC} & \multicolumn{2}{c|}{$ACC_b$ (before \OURMETHOD)}                & \multicolumn{2}{c|}{$ASR_b$ (before \OURMETHOD)}                     & \multicolumn{2}{c|}{$ACC_s$ (after \OURMETHOD)}          & \multicolumn{2}{c|}{$ASR_s$ (after \OURMETHOD)}    & \multicolumn{2}{c|}{$FID$}           \\ \cline{4-13} 
                         &                        &                           & \multicolumn{1}{c|}{Rf}      & \multicolumn{1}{c|}{Tj} &  \multicolumn{1}{c|}{Rf}      & \multicolumn{1}{c|}{Tj} & \multicolumn{1}{c|}{Rf} & \multicolumn{1}{c|}{Tj} & \multicolumn{1}{c|}{Rf} & \multicolumn{1}{c|}{Tj} & \multicolumn{1}{c|}{Rf} & \multicolumn{1}{c|}{Tj} \\ \hline
\multirow{4}{*}{MNIST}   & ShuffleNetx1.0         & 99.14\%            &        
                        \multicolumn{1}{c|}{99.16\%} & \multicolumn{1}{c|}{99.72\%}   & \multicolumn{1}{c|}{100\%}   & \multicolumn{1}{c|}{100\%}   & \multicolumn{1}{c|}{98.05\%}   & \multicolumn{1}{c|}{97.28\%}   & \multicolumn{1}{c|}{0.91\%}   & \multicolumn{1}{c|}{0\%}   & \multicolumn{1}{c|}{97.96\%} & \multicolumn{1}{c|}{97.55\%} \\ \cline{2-13} 
                         & Vgg16                  & 99.38\%                   & 
                         \multicolumn{1}{c|}{99.37\%} & \multicolumn{1}{c|}{99.00\%}   &   \multicolumn{1}{c|}{100\%}   & \multicolumn{1}{c|}{100\%}   &  \multicolumn{1}{c|}{97.07\%}   & \multicolumn{1}{c|}{97.03\%}   &   \multicolumn{1}{c|}{0.78\%}   & \multicolumn{1}{c|}{0\%}   &
                         \multicolumn{1}{c|}{96.90\%}   & \multicolumn{1}{c|}{98.01\%}  \\ \cline{2-13} 
                         & ResNet18               & 99.69\%                   &
                         \multicolumn{1}{c|}{99.35\%} & \multicolumn{1}{c|}{98.57\%}     & \multicolumn{1}{c|}{100\%}   & \multicolumn{1}{c|}{100\%}     & \multicolumn{1}{c|}{98.30\%}   & \multicolumn{1}{c|}{98.21\%}     & \multicolumn{1}{c|}{0.76\%}   & \multicolumn{1}{c|}{0\%}     &
                         \multicolumn{1}{c|}{98.18\%}   & \multicolumn{1}{c|}{99.63\%}     \\ \cline{2-13} 
                         & ResNet101              & 99.63\%                   & 
                         \multicolumn{1}{c|}{98.39\%}        & \multicolumn{1}{c|}{98.20\%}     & \multicolumn{1}{c|}{100\%}        & \multicolumn{1}{c|}{100\%}    & \multicolumn{1}{c|}{97.88\%}   & \multicolumn{1}{c|}{97.52\%}     & \multicolumn{1}{c|}{0.83\%}   & \multicolumn{1}{c|}{0\%}     &
                         \multicolumn{1}{c|}{98.64\%}   & \multicolumn{1}{c|}{99.31\%}     \\ 
                         \hline
\multirow{4}{*}{GTSRB}   & ShuffleNetx1.0         & 99.72\%     &
                         \multicolumn{1}{c|}{97.07\%}        & \multicolumn{1}{c|}{99.78\%}    & \multicolumn{1}{c|}{99.68\%}        & \multicolumn{1}{c|}{100\%}     & \multicolumn{1}{c|}{95.03\%}   & \multicolumn{1}{c|}{97.57\%}     & \multicolumn{1}{c|}{0.66\%}   & \multicolumn{1}{c|}{0.71\%}     &
                         \multicolumn{1}{c|}{97.22\%}   & \multicolumn{1}{c|}{97.07\%}     \\\cline{2-13} 
                         & Vgg16                  & 97.67\%                   & 
                         \multicolumn{1}{c|}{94.70\%} & \multicolumn{1}{c|}{98.37\%}     & \multicolumn{1}{c|}{99.98\%} & \multicolumn{1}{c|}{100\%}     & \multicolumn{1}{c|}{92.97\%}   & \multicolumn{1}{c|}{96.34\%}     & \multicolumn{1}{c|}{0.81\%}   & \multicolumn{1}{c|}{0.66\%}     &
                         \multicolumn{1}{c|}{97.32\%}   & \multicolumn{1}{c|}{97.27\%}     \\ \cline{2-13} 
                         & ResNet18               & 99.85\%                   & 
                         \multicolumn{1}{c|}{94.29\%} & \multicolumn{1}{c|}{98.56\%}     & \multicolumn{1}{c|}{99.98\%}   & \multicolumn{1}{c|}{100\%}     & \multicolumn{1}{c|}{93.86\%}   & \multicolumn{1}{c|}{97.21\%}     & \multicolumn{1}{c|}{0.75\%}   & \multicolumn{1}{c|}{0.63\%}    &
                         \multicolumn{1}{c|}{98.75\%}   & \multicolumn{1}{c|}{97.99\%}    \\ \cline{2-13} 
                         & ResNet101              & 99.83\%                   & 
                         \multicolumn{1}{c|}{97.94\%}        & \multicolumn{1}{c|}{98.33\%}     & \multicolumn{1}{c|}{100\%}        & \multicolumn{1}{c|}{100\%}    & \multicolumn{1}{c|}{95.64\%}   & \multicolumn{1}{c|}{96.98\%}     & \multicolumn{1}{c|}{0.92\%}   & \multicolumn{1}{c|}{0.89\%}     &
                         \multicolumn{1}{c|}{96.71\%}   & \multicolumn{1}{c|}{97.72\%}     \\ \hline
\multirow{4}{*}{CIFAR10} & ShuffleNetx1.0         & 94.63\%                   & 
                        \multicolumn{1}{c|}{90.60\%}        & \multicolumn{1}{c|}{94.41\%}     & \multicolumn{1}{c|}{100\%}        & \multicolumn{1}{c|}{100\%}    & \multicolumn{1}{c|}{90.02\%}   & \multicolumn{1}{c|}{92.64\%}    & \multicolumn{1}{c|}{1.57\%}   & \multicolumn{1}{c|}{2.34\%}     &
                        \multicolumn{1}{c|}{97.63\%}   & \multicolumn{1}{c|}{95.65\%}     \\ \cline{2-13} 
                         & Vgg16                  & 95.12\%                   & 
                         \multicolumn{1}{c|}{91.62\%} & \multicolumn{1}{c|}{95.11\%}     & \multicolumn{1}{c|}{99.95\%} & \multicolumn{1}{c|}{100\%}     & \multicolumn{1}{c|}{90.99\%}   & \multicolumn{1}{c|}{94.25\%}     & \multicolumn{1}{c|}{1.17\%}   & \multicolumn{1}{c|}{3.67\%}  &
                         \multicolumn{1}{c|}{98.04\%}   & \multicolumn{1}{c|}{95.24\%}  \\ \cline{2-13} 
                         & ResNet18               & 96.50\%                   & 
                         \multicolumn{1}{c|}{93.09\%} & \multicolumn{1}{c|}{96.50\%}      & \multicolumn{1}{c|}{100\%}   & \multicolumn{1}{c|}{100\%}    & \multicolumn{1}{c|}{92.07\%}   & \multicolumn{1}{c|}{96.01\%}     & \multicolumn{1}{c|}{2.10\%}   & \multicolumn{1}{c|}{3.17\%}   &
                         \multicolumn{1}{c|}{96.65\%}   & \multicolumn{1}{c|}{96.21\%}   \\ \cline{2-13} 
                         & ResNet101              & 96.98\%                   & 
                         \multicolumn{1}{c|}{91.24\%} & \multicolumn{1}{c|}{96.98\%}     & \multicolumn{1}{c|}{100\%}   & \multicolumn{1}{c|}{100\%}    & \multicolumn{1}{c|}{90.79\%}   & \multicolumn{1}{c|}{95.82\%}      & \multicolumn{1}{c|}{2.22\%}   & \multicolumn{1}{c|}{2.55\%}    &
                         \multicolumn{1}{c|}{97.07\%}   & \multicolumn{1}{c|}{96.17\%}    \\ \hline
\end{tabular}
\end{adjustbox}
\end{table*}

\begin{table*}[htp]
\centering
\caption{Comparison of \OURMETHOD with backdoor defenses on MNIST, GTSRB and CIFAR10. NC represents Neural Cleanse defense, FP represents Fine Pruning defense. \OURMETHOD forgets on 0.1\% training data and recovers on 10\% training data.}
\label{tab:evaluation_defenses}
\begin{adjustbox}{width=0.80\textwidth}
\begin{tabular}{|l|c|clllll|llllll|}
\hline
\multicolumn{1}{|c|}{\multirow{3}{*}{Dataset}} & \multirow{3}{*}{Model} & \multicolumn{6}{c|}{Reflection}                                                                                                                    & \multicolumn{6}{c|}{TrojanNet}                                                                                                              \\ \cline{3-14} 
\multicolumn{1}{|c|}{}                         &                        & \multicolumn{3}{c|}{\textit{Fidelity} ($FID$)}                                                   & \multicolumn{3}{c|}{Time (seconds)}                        & \multicolumn{3}{c|}{\textit{Fidelity} ($FID$)}                                              & \multicolumn{3}{c|}{Time (seconds)}                      \\ \cline{3-14} 
\multicolumn{1}{|c|}{}                         &                        & \multicolumn{1}{c|}{NC}      & \multicolumn{1}{c|}{FP} & \multicolumn{1}{c|}{SEAM}    & \multicolumn{1}{c|}{NC}   & \multicolumn{1}{c|}{FP} & SEAM & \multicolumn{1}{c|}{NC} & \multicolumn{1}{c|}{FP} & \multicolumn{1}{c|}{SEAM}    & \multicolumn{1}{c|}{NC} & \multicolumn{1}{c|}{FP} & SEAM \\ \hline
\multirow{4}{*}{MNIST}  & ShuffleNetx1.0         & 
                        \multicolumn{1}{c|}{14.64\%}        & \multicolumn{1}{c|}{89.50\%}   & \multicolumn{1}{c|}{97.96\%} & \multicolumn{1}{c|}{825}     & \multicolumn{1}{c|}{575}   &   \multicolumn{1}{c|}{104}   & 
                        \multicolumn{1}{c|}{100\%}   & \multicolumn{1}{c|}{89.13\%}   & \multicolumn{1}{c|}{97.55\%} & \multicolumn{1}{c|}{818}   & \multicolumn{1}{c|}{632}   &  \multicolumn{1}{c|}{112}    \\ \cline{2-14} 
                        & Vgg16                  & \multicolumn{1}{c|}{$< 0\%$ }        & \multicolumn{1}{c|}{87.95\%}   & \multicolumn{1}{c|}{96.90\%} &  \multicolumn{1}{c|}{793}     & \multicolumn{1}{c|}{721}   &  \multicolumn{1}{c|}{229}     & \multicolumn{1}{c|}{90.03\%}   & \multicolumn{1}{c|}{89.22\%}   & \multicolumn{1}{c|}{98.01\%} & \multicolumn{1}{c|}{802}   & \multicolumn{1}{c|}{701}   &   \multicolumn{1}{c|}{225}   \\ \cline{2-14} 
                        & ResNet18               & \multicolumn{1}{c|}{72.75\%}        & \multicolumn{1}{c|}{90.17\%}   & \multicolumn{1}{c|}{98.18\%} & \multicolumn{1}{c|}{1163}     & \multicolumn{1}{c|}{909}   &  \multicolumn{1}{c|}{329}    & \multicolumn{1}{c|}{100\%}   & \multicolumn{1}{c|}{100.97\%}   & \multicolumn{1}{c|}{99.63\%} & \multicolumn{1}{c|}{1148}   & \multicolumn{1}{c|}{938}   &  \multicolumn{1}{c|}{341}     \\ \cline{2-14} 
                        & ResNet101              & \multicolumn{1}{c|}{$<$ 0\%}        & \multicolumn{1}{c|}{90.89\%}   & \multicolumn{1}{c|}{98.64\%} & \multicolumn{1}{c|}{7865}     & \multicolumn{1}{c|}{4445}   & \multicolumn{1}{c|}{935}      & \multicolumn{1}{c|}{89.61\%}   & \multicolumn{1}{c|}{90.27\%}   & \multicolumn{1}{c|}{99.31\%} & \multicolumn{1}{c|}{7361}   & \multicolumn{1}{c|}{4415}   & \multicolumn{1}{c|}{1018}      \\ \hline
\multirow{4}{*}{GTSRB}                         
                        & ShuffleNetx1.0         & \multicolumn{1}{l|}{96.72\%} & \multicolumn{1}{c|}{99.65\%}   & \multicolumn{1}{c|}{97.22\%} & \multicolumn{1}{c|}{3442} & \multicolumn{1}{c|}{1040}   &  \multicolumn{1}{c|}{218}    & \multicolumn{1}{c|}{99.97\%}   & \multicolumn{1}{c|}{86.35\%}   & \multicolumn{1}{c|}{97.07\%} & \multicolumn{1}{c|}{3648}   & \multicolumn{1}{c|}{770}   &     \multicolumn{1}{c|}{217}  \\ \cline{2-14} 
                        & Vgg16                  & \multicolumn{1}{c|}{96.36\%} & \multicolumn{1}{c|}{100.25\%}   & \multicolumn{1}{c|}{97.32\%} & \multicolumn{1}{c|}{2634} & \multicolumn{1}{c|}{909}   &    \multicolumn{1}{c|}{413}   & \multicolumn{1}{c|}{99.41\%}   & \multicolumn{1}{c|}{96.05\%}   & \multicolumn{1}{c|}{97.27\%} & \multicolumn{1}{c|}{2735}   & \multicolumn{1}{c|}{697}   &   \multicolumn{1}{c|}{423}   \\ \cline{2-14} 
                        & ResNet18               & \multicolumn{1}{c|}{98.49\%} & \multicolumn{1}{c|}{99.98\%}   & \multicolumn{1}{c|}{98.75\%} & \multicolumn{1}{c|}{3424} & \multicolumn{1}{c|}{792}   &   \multicolumn{1}{c|}{601}    & \multicolumn{1}{c|}{99.43\%}   & \multicolumn{1}{c|}{93.95\%}   & \multicolumn{1}{c|}{97.99\%} & \multicolumn{1}{c|}{3523}   & \multicolumn{1}{c|}{874}   &  \multicolumn{1}{c|}{639}    \\ \cline{2-14} 
                        & ResNet101              & \multicolumn{1}{c|}{99.04\%}        & \multicolumn{1}{c|}{99.48\%}   & \multicolumn{1}{c|}{96.71\%} & \multicolumn{1}{c|}{13287}     & \multicolumn{1}{c|}{4990}   &  \multicolumn{1}{c|}{2052}    & \multicolumn{1}{c|}{99.47\%}   & \multicolumn{1}{c|}{97.28\%}   & \multicolumn{1}{c|}{97.72\%} & \multicolumn{1}{c|}{13948}   & \multicolumn{1}{c|}{5000}   &   \multicolumn{1}{c|}{2412}    \\ \hline
\multirow{4}{*}{CIFAR10}                       
                        & ShuffleNetx1.0         & \multicolumn{1}{c|}{90.79\%} & \multicolumn{1}{c|}{82.03\%}   & \multicolumn{1}{c|}{97.63\%} & \multicolumn{1}{c|}{1677} & \multicolumn{1}{c|}{1033}   &   \multicolumn{1}{c|}{231}    & \multicolumn{1}{c|}{89.23\%}   & \multicolumn{1}{c|}{86.95\%}   & \multicolumn{1}{c|}{96.15\%} & \multicolumn{1}{c|}{1725}   & \multicolumn{1}{c|}{1131}   &  \multicolumn{1}{c|}{253}     \\ \cline{2-14} 
                        & Vgg16                  & \multicolumn{1}{c|}{57.17\%} & \multicolumn{1}{c|}{85.24\%}   & \multicolumn{1}{c|}{98.04\%} & \multicolumn{1}{c|}{1232} & \multicolumn{1}{c|}{966}   &    \multicolumn{1}{c|}{425}   & \multicolumn{1}{c|}{89.23\%}   & \multicolumn{1}{c|}{88.36\%}   & \multicolumn{1}{c|}{95.24\%} & \multicolumn{1}{c|}{1277}   & \multicolumn{1}{c|}{987}   &   \multicolumn{1}{c|}{443}    \\ \cline{2-14} 
                        & ResNet18               & \multicolumn{1}{c|}{88.96\%} & \multicolumn{1}{c|}{85.26\%}   & \multicolumn{1}{c|}{96.65\%} & \multicolumn{1}{c|}{1662} & \multicolumn{1}{c|}{1293}   &  \multicolumn{1}{c|}{686}     & \multicolumn{1}{c|}{89.44\%}   & \multicolumn{1}{c|}{97.19\%}   & \multicolumn{1}{c|}{96.15\%} & \multicolumn{1}{c|}{1695}   & \multicolumn{1}{c|}{1404}   &   \multicolumn{1}{c|}{679}    \\ \cline{2-14} 
                        & ResNet101              & \multicolumn{1}{c|}{89.52\%}        & \multicolumn{1}{c|}{92.25\%}   & \multicolumn{1}{c|}{97.07\%} & \multicolumn{1}{c|}{7763}     & \multicolumn{1}{c|}{5114}   &   \multicolumn{1}{c|}{2910}     & \multicolumn{1}{c|}{89.67\%}   & \multicolumn{1}{c|}{95.19\%}   & \multicolumn{1}{c|}{96.17\%} & \multicolumn{1}{c|}{7277}   & \multicolumn{1}{c|}{5283}   &  \multicolumn{1}{c|}{3008}     \\ \hline
\end{tabular}
\end{adjustbox}
\vspace{-1em}
\end{table*}

\vspace{3pt}\noindent\textbf{On NLP tasks}.
We evaluated \OURMETHOD on NLP tasks using datasets of \textit{Round 5-7} of the TrojAI competition. These sentiment classification models are built by stacking a classification model on top of pre-trained transformers~\cite{DBLP:conf/emnlp/WolfDSCDMCRLFDS20} (e.g., BERT). During training, the transformers are fixed and only the weights of the classification models are updated, which mimics the popular pipeline of NLP tasks. In this way, the backdoor will only affect the classification model. The classification models in the datasets have three different architectures: GRU~\cite{GRU}, LSTM~\cite{LSTM} and Fully Connected (FC) networks, and use various hyper-parameters (e.g., different number of layers). Models of \textit{Round 7} are trained for NER and built by stacking a linear layer upon 4 kinds of transformers (Section~\ref{subsec:datasets}). During the training, the weights of both the transformers and the linear layers are updated.

\vspace{-3pt}

In our experiments, again, we utilized the small set of clean data ($0.1\%$ of the training set) provided by the competition organizer for the \textit{forgetting} and the \textit{recovery} steps; we also followed the competition's documents~\cite{iarpa} to generate the testing data. Table~\ref{tab:evaluation_trojai} illustrates the experimental results. As we can see from the table, \OURMETHOD achieves an average FID of $88.30\%$,  $89.16\%$ and $92.65\%$ on the datasets of \textit{Round 5} and \textit{Round 6} and \textit{Round 7} respectively. The relatively higher \textit{Fidelity} in the last round could be attributed to the higher ACC of the infected models in the round ($93.60\%$ vs. an average $90.15\%$ of the models in other rounds), and the availability of more clean training data: 
in the sentiment analysis (\textit{Round 5-6}), one clean sentence released by the competition organizer can only be used as one instance in the training set (for both the \textit{forgetting} and \textit{recovery} operations), while the same sentence can be broken down to multiple entity-related terms for unlearning backdoors in an NER model, the \textit{Round 7} task. 

\vspace{3pt}\noindent\textbf{On clean model.} To understand the impacts of \OURMETHOD on the accuracy of the clean model, we performed experiments on CIFAR10, with the results reported in Table~\ref{clean_model_table}, i.e., the change of the clean models' ACC before and after running SEAM. In the experiments, we again utilized a small set of clean data ($0.1\%$ of the training set) for the \textit{forgetting} and the \textit{recovery} steps. As we can see from Table 3, an average accuracy loss caused by \OURMETHOD is just $1\%$ on the clean models with four mainstream structures.

\begin{table}[htp] 
\centering
\caption{The task accuracies of the clean models before and after processed by \OURMETHOD}
\label{clean_model_table}
\begin{adjustbox}{width=0.2\textwidth}
\begin{tabular}{|l|l|l|}
\hline & \text { Before } & \text { After } \\
\hline \text { ShuffleNetx1.0 } & $94.63 \%$ & $93.14 \%$ \\
\hline \text { Vgg16 } & $95.12 \%$ & $94.42 \%$ \\
\hline \text { ResNet18 } & $93.09 \%$ & $92.18 \%$ \\
\hline \text { ResNet101 } & $96.98 \%$ & $95.73 \%$ \\
\hline
\end{tabular}
\end{adjustbox}
\end{table}
\vspace{-3mm}

\begin{table}[htp] \label{table5}
\centering
\caption{Comparison of \OURMETHOD with backdoor defenses on Trojai competition \textit{Round 3-7}. NC represents Neural Cleanse defense, FP represents Fine Pruning defense. Results are averaged among all backdoor infected models in each round. \OURMETHOD forgets and recovers on the clean dataset provided for each model (about $0.1\%$ of the training dataset). The results are averaged among all models for each round. The unit of time is second.}
\label{tab:evaluation_trojai}
\begin{adjustbox}{width=0.40\textwidth}
\begin{tabular}{|cl|l|l|l|l|l|}
\hline
\multicolumn{2}{|l|}{}                             & \multicolumn{1}{c|}{Round 3} & \multicolumn{1}{c|}{Round 4} & \multicolumn{1}{c|}{Round 5} & \multicolumn{1}{c|}{Round 6} & \multicolumn{1}{c|}{Round 7} \\ \hline
\multicolumn{1}{|c|}{\multirow{3}{*}{FID}}  & NC   &    53.42\% &  54.15\%       &    67.27\%                          &           64.18\%                   &           61.90\%                   \\ \cline{2-7} 
\multicolumn{1}{|c|}{}                      & FP   &     71.33\%  &   72.81\%  &   64.49\%  &   63.39\%  & 67.13\%     \\ \cline{2-7} 
\multicolumn{1}{|c|}{}                      & SEAM & 89.84\%                      & 88.75\%                      & 88.30\%                      & 89.16\%                      & 92.65\%                      \\ \hline
\multicolumn{1}{|c|}{\multirow{3}{*}{Time}} & NC   &        5218           &            4733                  &           2763                   &          2893                    &               4919               \\ \cline{2-7} 
\multicolumn{1}{|c|}{}                      & FP   &   2107    &     2053     &     2378           &  2439      &     2643        \\ \cline{2-7} 
\multicolumn{1}{|c|}{}                      & SEAM & 39                           & 39                           & 41                           & 47                           & 49                           \\ \hline
\end{tabular}
\end{adjustbox}
\vspace{-1.5em}
\end{table}

\vspace{-1mm}
\subsection{Efficiency}
\vspace{-4mm}
\label{subsec:efficiency}

We further analyzed the efficiency of \OURMETHOD from two aspects: execution time and clean data size. 

\noindent\textbf{Execution time}. 
Overall, \OURMETHOD is found to be highly efficient, vastly outperforming other unlearning techniques (see Table~\ref{tab:evaluation_defenses} and~\ref{tab:evaluation_trojai}) in execution time. Particularly, from Table~\ref{tab:evaluation_defenses}, we can see that on various models trained on the three popular datasets (MNIST/GTSRB/CIFAR10),  \OURMETHOD typically just needs no more than 12 minutes to nearly completely remove the backdoor effect. The only exception is ResNet101, which uses massive GPU memory, so we had to reduce its batch size to 8, instead of 32 set for other model architectures.  


Then we take a close look at the \textit{forgetting} and \textit{recovery} steps. The time complexities of these steps are $\mathcal{O}(\mathbf{N}_{for})$ and $\mathcal{O}(\mathbf{N}_{rec})$, where the former is the number of epochs for training the random-labeling task of the \textit{forgetting} step, and the latter is the number of epochs for training the primary task for the \textit{recovery} step. We compared the execution time of SEAM on infected ResNet18 models on three datasets (MNIST, GTSRB and CIFAR10) with these models' original training time. As we can see from the results in Table~\ref{tab:train_from_scratch}, on average, SEAM takes less than 7\% of the time for training a model from scratch to suppress the model's backdoor effect while preserving its legitimate classification capability.

We found that, on MNIST/GTSRB/CIFAR10, the \textit{forgetting} step takes much less time than the \textit{recovery} step, since 1) the \textit{forgetting} typically needs less than 10 epochs while the \textit{recovery} requires at most 100 epochs, and 2) the dataset for the \textit{forgetting} is just $0.1\%$ of the training dataset, while the dataset for the \textit{recovery} was set to $10\%$ in our experiments on MNIST/GTSRB/CIFAR10. On \textit{Round 3-7} of TrojAI competition, the \textit{recovery} dataset is only $0.1\%$ training dataset that largely accelerated the \textit{recovery} step with the cost of lightly reduced \textit{Fidelity}. The effect of clean data size will be evaluated later.

\begin{figure}[h]
\vspace{-1.0em}
\centering
\includegraphics[width=0.3\textwidth]{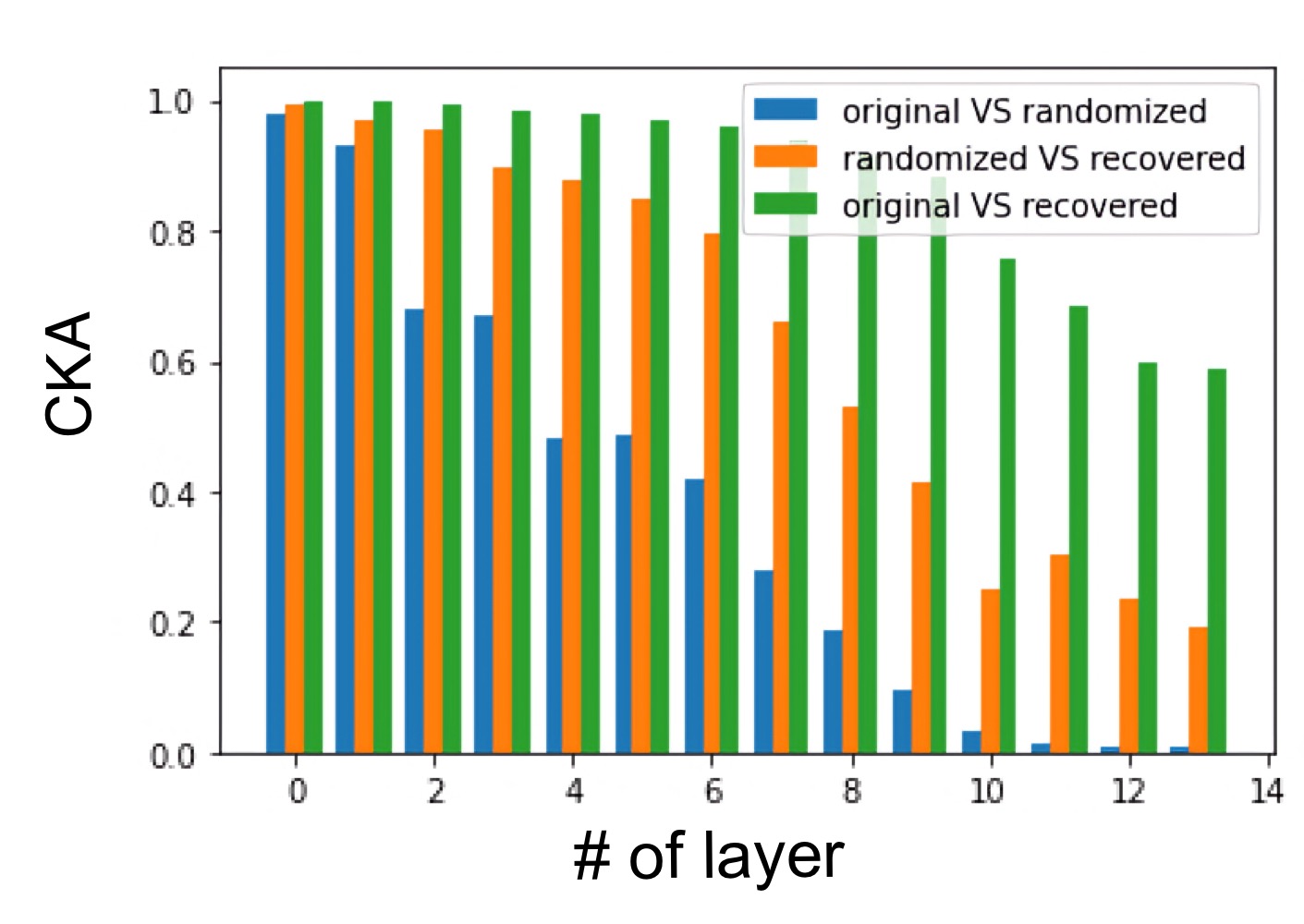}
\caption{CKA on each layer of VGG16. Layer 0 is the first layer (input) and layer 14 is the last second layer.}
\label{fig:CKA}
\vspace{-1em}
\end{figure}

To understand what makes \OURMETHOD so efficient, we looked into the changes of each layer after the \textit{forgetting} step and the \textit{recovery} step. Specifically, for a VGG16 model, we measured the similarity of the same layer between the original model (the backdoored one before unlearning), the randomized model (after the \textit{forgetting} step) and the recovered model (after the \textit{recovery} step). Here we use the centered kernel alignment (CKA)~\cite{CKApaper} as the metric, which ranges in $[0,1]$ with $CKA=1$ being identical and $CKA=0$ being totally different. Fig.~\ref{fig:CKA} demonstrates the CKA results for a VGG16 model on a clean dataset. As we can see here, for each layer, the closer it is located to the output, the more different can been seen between the original model and the randomized model. This indicates that many features of the original model has been preserved during the \textit{forgetting} step, particularly those on the shallow layers of the model, so the \textit{recovery} step only needs to restore the features on the layers more toward the output layer, thereby allowing a faster unlearning and recovering.



\vspace{3pt}\noindent\textbf{Clean data size}.
As mentioned earlier, we used $10\%$ of the training data for recovering on MNIST/GTSRB/CIFAR10, and only $0.1\%$ of the training data on \textit{Round 3-7} datasets of the TrajAI competition. 
As a result, \OURMETHOD runs much faster on the \textit{Round 3-7} datasets, less than one minute on average for each model, though the \textit{Fidelity} of the unlearning goes down a little bit ($>88\%$ vs. $>95\%$).

\begin{figure}[h]
\vspace{-10pt}
\centering
\includegraphics[width=0.3\textwidth]{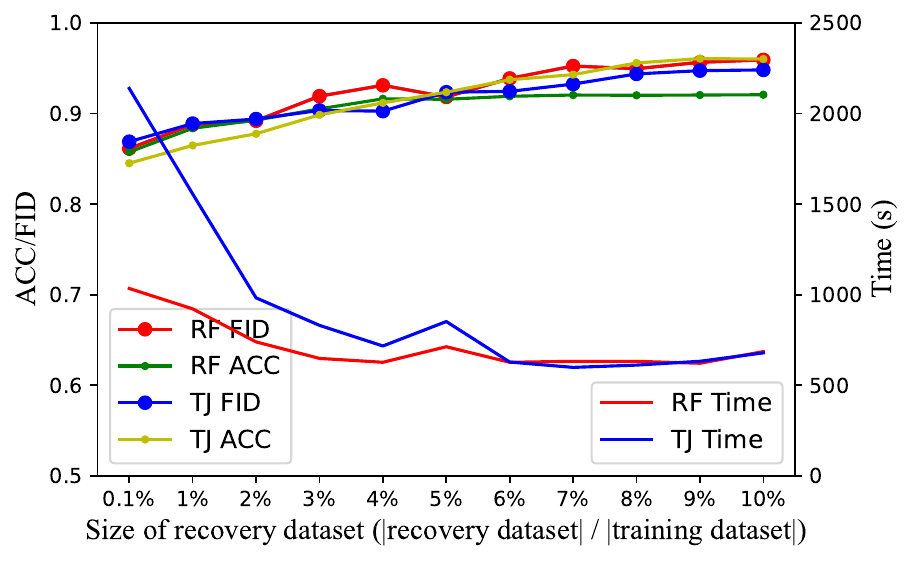}
\caption{Running time, \textit{Fidelity} (FID)  and accuracy (ACC) of the recovered models. Here RF represents the effects of SEAM against the Reflection attack, and TJ represents the effects of SEAM against the TrojanNet attack.}
\label{fig:fed_trend}
\vspace{-1em}
\end{figure}

To understand the impact of the size of \textit{recovery} datasets $\mathcal{D}_{rec}$ on the effectiveness of unlearning, we ran \OURMETHOD on $\mathcal{D}_{rec}$ of different sizes (randomly drawn from CIFAR10). The results are presented in Fig.~\ref{fig:fed_trend}. We observe that the execution time of \OURMETHOD goes down  along with the increase of the size of $\mathcal{D}_{rec}$ against Reflection attacks and TrojanNet attack.
while the \textit{Fidelity} changes gently. 
Again, we confirm that indeed \OURMETHOD only needs an exceedingly small amount of clean data ($0.1\%$, about 5 images per class) to achieve a decent unlearning effect (\textit{Fidelity} of $86\%$).




\vspace{-3mm}
\subsection{Comparison}
\label{subsec:comparison}
\vspace{-4mm}

We compared \OURMETHOD with Neural Cleanse (NC)~\cite{NeuralCleanse}, Fine-Pruning (FP)~\cite{FinePruning}, fine-tuning, naive continuous-training, and Neural Attention Distillation (NAD)~\cite{NAD}, five representative unlearning techniques. 
As mentioned earlier, NC is a detection-based approach that first recovers the trigger from a backdoored model and then removes the backdoor from the model through unlearning (i.e., retraining the model on the trigger-carrying inputs with the correct labels). FP is a blind unlearning approach that prunes a backdoored model and then fine-tunes it on a clean dataset, in an attempt to remove the backdoor effect. Fine-tuning is a method that tunes the parameters of a backdoored model's last two layers on a small set of clean data using gradient descent; continuous-training keeps on training the model on a small set of clean data, hoping to weaken its backdoor effect when it is infected. NAD is a blind unlearning approach that uses a clean dataset to distillate a new clean model from the victim model.

For a fair comparison, we implemented \OURMETHOD under the TrojanZoo framework~\cite{TrojanZoo} and utilized the implementations of NC and FP provided by the TrojanZoo team. We also kept the unlearning datasets for NC, FP and \OURMETHOD \ignore{ (\textit{recovery} step) be}to the same size, i.e., $10\%$ of the training datasets for MNIST/GTSRB/CIFAR10 and $0.1\%$ for the TrojAI competition. Also, for Fine-pruning, we followed the FP paper~\cite{FinePruning} to set the prune ratio to 0.82 of all neurons (i.e., maximum $82\%$ neurons would be pruned) and the fine-tuning epochs to 300 for MNIST, GTSRB and CIFAR10, 1000 for TrojAI competition \textit{Round 3-7}.

Table~\ref{tab:evaluation_defenses} and Table~\ref{tab:evaluation_trojai} show the comparison of \textit{Fidelity} among these solutions against the Reflection and TrojanNet attacks on MNIST/GTSRB/CIFAR10 and TrojAI datasets. We observe that \OURMETHOD achieves a high and stable \textit{Fidelity} against both attacks on all eight datasets,  while FP performs well against both attacks on MNIST and GTSRB, and NC only does well on GTSRB. The failure of NC against the Reflection attacks on MNIST and CIFAR10 could be attributed to its limitation in finding large triggers, as the Reflection attack may use an entire clean image as the trigger. The failure of FP against the Reflection attacks on CIFAR10 could be due to the difficulty in reducing ASR when trigger-relevant neurons have not been completely pruned. 

A larger \textit{Fidelity} gap between \OURMETHOD and NC/FP can be observed on \textit{Round 3-7} datasets of the TrojAI competition. The small clean dataset ($0.1\%$ of training data) available for unlearning significantly reduces the efficacy of NC/FP. Specifically, on \textit{Round 3-7}, the maximum \textit{Fidelity} achieved by NC is $67.27\%$ on the \textit{Round 5} data, and for FP, it is $72.81\%$ on the \textit{Round 4} data, which are far below the performance of \OURMETHOD on the same datasets: $88.30\%$ on \textit{Round 5} and $88.75\%$ on \textit{Round 4}.
Table~\ref{tab:evaluation_defenses} and Table~\ref{tab:evaluation_trojai} also compare the execution times of these solutions. On MNIST/GTSRB/CIFAR10, \OURMETHOD takes on average $1/6$ of the execution time used by NC and $1/4$ by FP. On \textit{Round 3-7} of the TrojAI competition, \OURMETHOD takes on average just $1\%$ of the execution time for NC and $1/50$ for FP. The high efficiency of \OURMETHOD can be attributed to the fact that the \textit{recovery} step only needs to restore the features on the layers close to the output (Section~\ref{subsec:efficacy}).

\OURMETHOD also significantly outperforms NAD and two simple baselines (fine-tuning and continuous training), especially when the clean data is scarce. In Table~\ref{NAD_table}, we present the \textit{Fidelity} results on the ResNet18 models for comparing SEAM, NAD and the two baselines on different sizes of clean datasets against the TrojanNet attack.

\begin{table}[htp] 
\vspace{-5pt}
\centering
\caption{The \textit{Fidelity} results of NAD, two base-line unlearning methods (Fine-tune and Continuous-training) and \OURMETHOD using the clean datasets of different sizes (compared with the training data size).}
\label{NAD_table}
\begin{adjustbox}{width=0.48\textwidth}
\begin{tabular}{|l|l|l|l|}
\hline & \text { MNIST } & \text { GTSRB } & \text { CIFAR10 } \\
\hline \text {SEAM (10\%  of training data size)} & 96.21 \% & 97.99 \% & 96.21 \% \\
\hline \text {NAD (10\% of training data size)} & 95.34 \% & 90.15 \% & 81.24 \% \\
\hline \text {Fine-tune (10\% of training data size)} & 51.12 \% & 47.32 \% & 44.29 \% \\
\hline \text {Continuous-training (10\% of training data size)} & 50.32 \% & 49.97 \% & 40.11 \% \\
\hline \text {SEAM (1\% of training data size)} & 96.31 \% & 94.72 \% & 89.04 \% \\
\hline \text {NAD (1\% of training data size)} & 64.57 \% & 59.35 \% & 56.35 \% \\
\hline \text {SEAM (0.1\% of training data size)} & 91.15 \% & 88.89 \% & 85.04 \% \\
\hline \text {NAD (0.1\% of training data size)} & 30.04 \% & 21.08 \% & 18.61 \% \\
\hline
\end{tabular}
\end{adjustbox}
\vspace{-0.5em}
\end{table}

As we can see from the table, NAD could not reasonably reduce ASR while maintaining a decent ACC when the size of the clean dataset goes down to 1\% and further to the 0.1\% of the training dataset size (note again that the clean data is not a subset of the training data). On the other hand, SEAM maintains its high effectiveness in unlearning, achieving 85\% to over 91\% \textit{Fidelity} with a small set of clean data (0.1\% of the training data size), in line with its performance on the TrojAI datasets, where for each model, only 10 samples are available for each class (Table~\ref{tab:evaluation_trojai}).





Compared with another naive baseline -- retraining the whole model from scratch on clean data, SEAM also demonstrates superior performance. We compared the performance of ResNet18 models recovered by SEAM and those trained from scratch on three datasets (MNIST, GTSRB and CIFAR10). As shown in Table~\ref{tab:train_from_scratch}, the models recovered by SEAM achieve on average 17\% higher ACC than the models trained from scratch on the same clean dataset (10\% of the whole training data size), and approach the ACC of the models trained from scratch on the whole training dataset. Also unlearning through SEAM takes about 45\% of the time for the training from scratch to converge on 10\% of the training data size (with much lower ACC) and only less than 7\% of the time for training on the whole dataset.

\begin{table}[htp] 
\centering
\caption{Comparison of $ACC$ and time cost between the ResNet18 models recovered by SEAM and those trained from scratch.}
\label{tab:train_from_scratch}
\begin{adjustbox}{width=0.48\textwidth}
\begin{tabular}{|c|c|c|c|c|}
\hline
                                                                                                            &      & MNIST      & GTSRB      & CIFAR10    \\ \hline
\multirow{2}{*}{\begin{tabular}[c]{@{}c@{}}Train from scratch \\ (same size as training data)\end{tabular}} & ACC  & 94.43\%    & 97.31\%    & 92.21\%    \\ \cline{2-5} 
                                                                                                            & Time & $\sim$3h   & $\sim$1.5h & $\sim$3h   \\ \hline
\multirow{2}{*}{\begin{tabular}[c]{@{}c@{}}Train from scratch \\ (10\% of training data size)\end{tabular}} & ACC  & 31.25\%    & 72.12\%    & 71.79\%    \\ \cline{2-5} 
                                                                                                            & Time & $\sim$0.4h & $\sim$0.3h & $\sim$0.4h \\ \hline
\multirow{2}{*}{\begin{tabular}[c]{@{}c@{}}SEAM\\ (10\% for recovery, 0.1\% for forgetting)\end{tabular}}   & ACC  & 92.79\%    & 97.16\%    & 91.43\%    \\ \cline{2-5} 
                                                                                                            & Time & $\sim$0.h  & $\sim$0.2h & $\sim$0.2h \\ \hline
\end{tabular}
\end{adjustbox}
\vspace{-0.5em}
\end{table}

\vspace{-3mm}
\subsection{Evasion}
\vspace{-4mm}
\label{subsec:evasion}

In this section, we investigate several possible evasion methods against SEAM, including the Label Consistent (LC) backdoor~\cite{label_consistent}, the Latent Separability (LS) backdoor, the Natural Backdoor (NB), the Entangled Watermarks (EW)~\cite{EW} and the evasion polluting the recover dataset with trigger-carrying inputs.

\begin{table}[htp]
\vspace{-6pt}
\centering
\caption{Effectiveness of \OURMETHOD against possible evasion methods on CIFAR10. $ACC_b$ and $ASR_b$ represents the ACC and ASR after attacks and before \OURMETHOD, $ACC_s$ and $ASR_s$ represents the ACC and ASR after \OURMETHOD, $FID$ represents the \textit{Fidelity} $\frac{ACC_s-ASR_s}{ACC_b}$. \OURMETHOD forgets on $0.1\%$ training data and recovers on $1\%$ training data. NB-t row represents results of \OURMETHOD forgets on $0.1\%$ training data and recovers on $10\%$ testing data. As the NB-t has same $ACC_b$, $ASR_b$, and very similar (less than 1\% difference) $ACC_s$ from NB, we only report $ASR_s$ for NB-t.}
\label{tab:evaluate_adaptive}
\begin{adjustbox}{width=0.35\textwidth}
\begin{tabular}{|c|c|c|c|c|c|}
\hline
\multicolumn{2}{|c|}{}                          & ShuffleNetx1.0 & Vgg16   & ResNet18 & ResNet101 \\ \hline
\multicolumn{1}{|c|}{\multirow{3}{*}{$ACC_b$}} 
                                        & LC & 90.54\%        & 91.28\% & 93.82\%  & 92.74\%   \\ \cline{2-6} 
\multicolumn{1}{|c|}{}                  & LS & 90.14\%        & 90.86\% & 91.61\%  & 92.11 \%  \\ \cline{2-6}
\multicolumn{1}{|c|}{}                  & NB & 94.63\%        & 95.12\% & 96.50\%  & 96.98\%   \\ \hline
\multicolumn{1}{|c|}{\multirow{3}{*}{$ASR_b$}} 
                                        & LC &  98.06\%        & 99.87\% & 99.78\%  & 87.70\%   \\ \cline{2-6} 
\multicolumn{1}{|c|}{}                  & LS &  98.17\%        & 98.68\% & 99.12 \% & 97.83 \%  \\ \cline{2-6} 
\multicolumn{1}{|c|}{}                  & NB &  84.73\%        & 76.59\% & 83.24\%  & 71.93\%   \\ \hline
\multicolumn{1}{|c|}{\multirow{3}{*}{$ACC_s$}} 
                                        & LC & 91.21\%        & 91.03\% & 94.18\%  & 92.53\%    \\ \cline{2-6} 
\multicolumn{1}{|c|}{}                  & LS & 90.02\%        & 90.13\% & 90.93\%  & 91.05 \% \\ \cline{2-6}
\multicolumn{1}{|c|}{}                  & NB &  94.27\%       &  95.18\%  & 95.64\%   & 91.99\%    \\ \hline
\multicolumn{1}{|c|}{\multirow{4}{*}{$ASR_s$}} 
                                        & LC &   7.19\%         & 7.93\%  &  9.21\%  &  9.34\%   \\ \cline{2-6} 
\multicolumn{1}{|c|}{}                  & LS &  23.02\%         & 22.51\% & 21.56\%  & 22.41\%   \\ \cline{2-6}
\multicolumn{1}{|c|}{}                  & NB &  54.23\%        & 43.15\%  &  57.80\%   &  59.61\%  \\ \cline{2-6}
\multicolumn{1}{|c|}{}                  & NB-t &  33.79\%        & 28.73\%  &  31.28\%   &  34.40\%  \\ \hline
\multicolumn{1}{|c|}{\multirow{3}{*}{FID}} 
                                        & LC & 92.80\%        & 91.04\% &  90.57\% & 89.70\%  \\ \cline{2-6} 
\multicolumn{1}{|c|}{}                  & LS & 73.33\%        & 74.42\% & 75.72 \% & 74.52\% \\ \cline{2-6} 
\multicolumn{1}{|c|}{}                  & NB & 42.31\%        & 54.70\% & 39.21\% & 33.39\%  \\ \hline
\end{tabular}
\end{adjustbox}
\vspace{-2mm}
\end{table}

\noindent\textbf{Label Consistent backdoor}. 
Label Consistent (LC) is a data poisoning backdoor attack aiming to inject a targeted backdoor that makes the victim model misclassify samples in a specific source class to a target class. The idea is to use correctly labeled yet trigger-carrying images (which can escape human inspection) to cause the prediction of the target label to heavily rely on the triggers~\cite{label_consistent}, so the triggers can be used to induce misclassification.    


To find out the robustness of \OURMETHOD against LC, we performed a set of experiments on CIFAR10. Table~\ref{tab:evaluate_adaptive} shows that our approach successfully reduces the ASR from $>90\%$ to $<10\%$ and recovers the ACC to the level similar to that of the original model. In the meantime, indeed LC weakens the effectiveness of \OURMETHOD by causing a higher ASR after the \textit{recovery} step, compared with the results of running \OURMETHOD against the Reflection and TrojanNet attacks on CIFAR10 (Table~\ref{tab:evaluation_results}). Note that unlike other data poisoning attacks, LC requires information about the target model's representation space, which can only be estimated through model transferability.  It is still less clear how likely transferring a backdoor in this way could succeed.


\noindent\textbf{Latent Separability backdoor}. 
The Latent separability (LS) backdoor is an emerging attack that aims to build a backdoored model by producing indistinguishable representations in the latent space for trigger-carrying inputs and clean inputs. As a result, the backdoor task will behave similarly as the primary task, making it harder to unlearn the backdoor without affecting the primary task.  In particular, we evaluated SEAM against the Adaptive-Blend~\cite{latent_backdoor} attack on CIFAR10 with four typical model architectures (ShuffleNetx1.0, Vgg16, ResNet18, and ResNet101). The $LS$ rows in Table~\ref{tab:evaluate_adaptive} show the experimental results. We observe that, after being processed by SEAM, the backdoored models retain a high accuracy ($ACC_s \sim 90\%$) for their primary task but are significantly weakened in terms of their backdoor effects (with the Attack Success Rate $ASR_s \sim 22\%$). These results demonstrate the effectiveness of SEAM in defending against the attacks with the capability to undermine unlearning.


\noindent\textbf{Natural backdoor}. 
NB is the backdoor naturally generated during training, without the interference of a malicious party. It is introduced by the imperfection of the model architecture, the training process, or the training data, etc. Injection of NB is a process less stable than the poisoning attack: two independent training of the same model on the same dataset may lead to different NBs (with different triggers). 

To evaluate the performance of \OURMETHOD against NB, we performed experiments on CIFAR10. Specifically, we first recovered the NB of the target model through trigger inversion~\cite{NeuralCleanse}. 
Then, we ran \OURMETHOD on the target model using a forgetting dataset and a recovery dataset, with 0.1\% and 1\% of the training data considered to be ``clean'' (without the recovered trigger), 
respectively. The results are shown in Table~\ref{tab:evaluate_adaptive}. We observe that \OURMETHOD reduces the ASR of the NB, with the maximum reduction of 33.44\% on VGG16. To further weaken the effect of NB, 
we utilized $10\%$ of the testing dataset (which amounts to 1\% of the training data in size but has no overlap with its content), for recovery. Then, we leveraged the remaining $90\%$ of the testing data to measure the ACC and the ASR of the unlearned model.
The results are presented by the NB-t row in Table~\ref{tab:evaluate_adaptive}, which indicates that the use of the clean data not in the training set for recovery could be more effective in suppressing NB.

\begin{figure}[htbp]
   \begin{minipage}{0.23\textwidth}
     \centering
     \includegraphics[width=\linewidth]{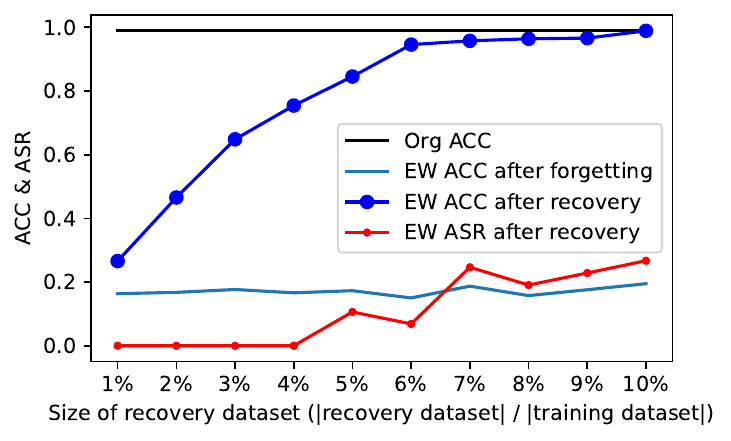}
     \caption{SEAM against EW on MNIST.}
     \label{fig:ew_mnist}
   \end{minipage}
   \hfill\noindent
   \begin{minipage}{0.23\textwidth}
     \centering
     \includegraphics[width=\linewidth]{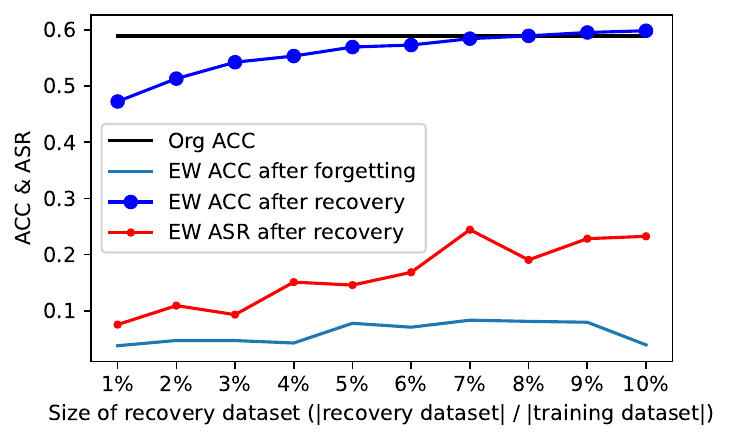}
     \caption{SEAM against EW on CIFAR100.}
     \label{fig:ew_cifar100}
   \end{minipage}
\vspace{-2mm}
\end{figure}

\noindent\textbf{Entangled watermarks}. 
EW injects a backdoor as a watermark for ownership protection into the target model. The idea is  to make the backdoor entangled with the primary task, so removal of the backdoor will undermine the target model's capability to perform its primary task. To understand whether SEAM still work on the models infected by EW, we conducted experiments over these models trained on the MNIST and CIFAR100 datasets.
Particularly, we utilized the source code of EW to generate the infected models and ran SEAM on them, using a clean dataset with the size of 0.1\% of the training data (for these models) for
forgetting and a clean dataset with the size of 1\%-10\% of the training data for recovery. Fig.~\ref{fig:ew_mnist} and Fig.~\ref{fig:ew_cifar100} show the results. As we can see from the figures, when the size of the recovery dataset is exceedingly small, noticeable performance degradation on benign inputs can still be observed from the infected models processed by SEAM, with the impact more conspicuous on the small task such as MNIST. Specifically, the recovered ACC becomes $\le 95\%$ when the size of the recovery dataset $\le 6\%$ of the training set on MNIST, and the ACC becomes $\le 54\%$ when the size of the recovery dataset $\le 5\%$ of the training set on CIFAR100. However, with moderate increase in the recovery data size, SEAM is found to be able to quickly restore the ACC: when the size of recovery dataset become $\ge 10\%$ ($\ge 8\%$) of the training set, the ACC of the recovered model gets back to or even goes beyond the ACC of the original model, that is, $99\%$ ($60\%$) on MNIST (CIFAR100). To further investigate how SEAM unlearns the EW-injected backdoor, we analyze the change of each layer within the target model under SEAM through a CKA experiment on CIFAR100 (see Fig.~\ref{fig:CKA_2} in  Appendix 3).

\vspace{-3mm}

\begin{figure}
\vspace{-4mm}
\centering
\includegraphics[width=0.23\textwidth]{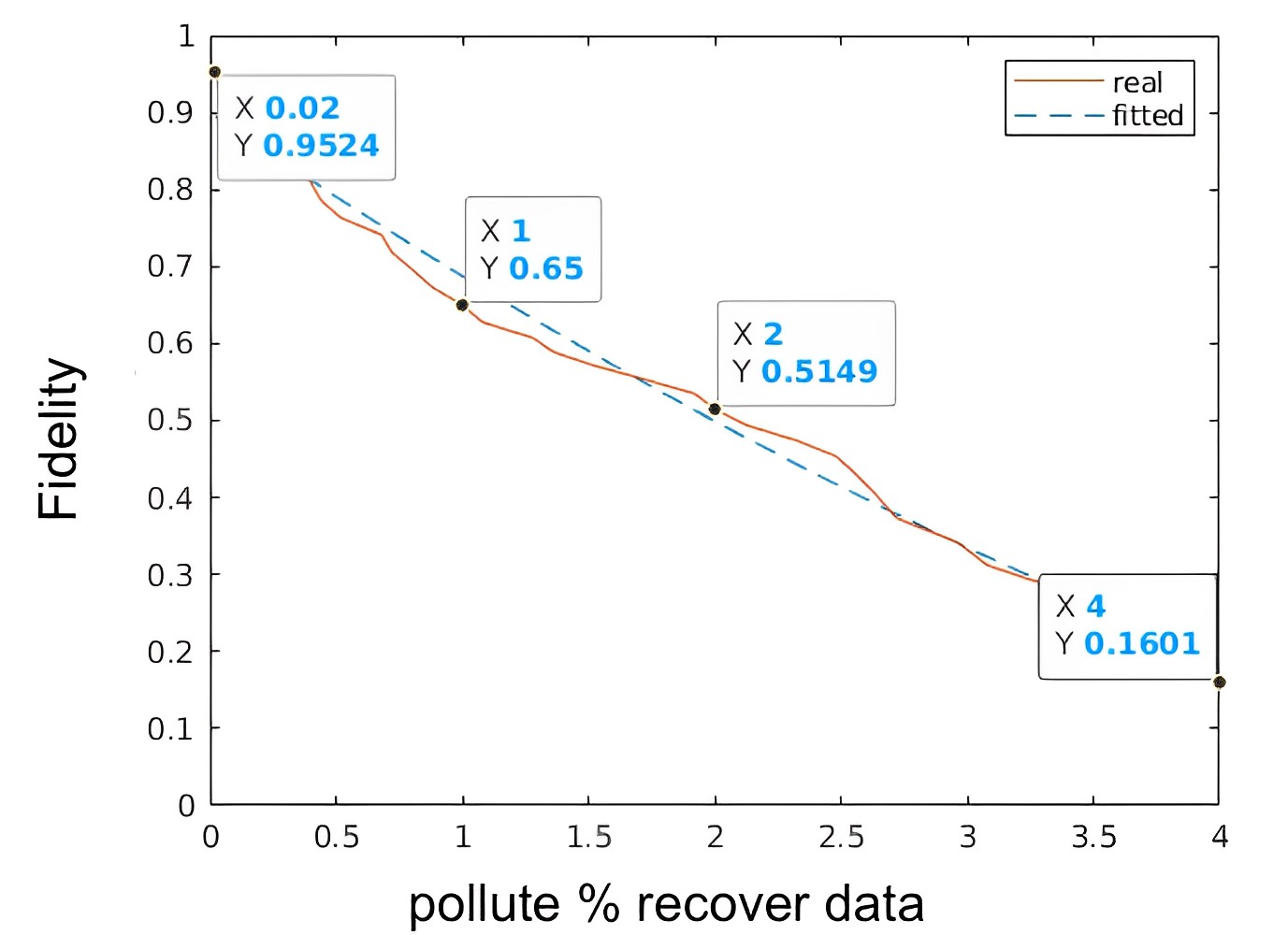}
\caption{The \textit{Fidelity} slowly decreases with the increasing trigger-carrying samples in the recovery dataset.}
\label{fig:few_backdoor}
\end{figure}

\noindent\textbf{Polluted recovery dataset}. 
To study the effect of data poisoning on SEAM's recovery step, we performed a k-out-of-n experiment, where we first polluted the recovery dataset so a portion of it are trigger-carrying inputs, according to a poisoning ratio, and then evaluated the impacts of the pollution on the protection of SEAM in terms of \textit{Fidelity}. From Fig.~\ref{fig:few_backdoor}, we observe that \textit{Fidelity} exhibits a negative correlation with the poisoning ratio of the recovery dataset. Specifically, when the poisoning ratio is 0.02\% (the recovery dataset including only 1 trigger-carrying input), the average \textit{Fidelity} achieved by SEAM on the models decreases from 98.2\% to 95.2\%. When the poisoning ratio becomes 1\% (i.e., the recovery dataset containing 50 trigger-carrying inputs), the \textit{Fidelity} further decreases to 65\%. These results demonstrate that SEAM has some resilience to data poisoning in the recovery dataset.



\vspace{-3mm}
\section{Related Work}\label{Related Works_section}
\vspace{-4mm}
There are mainly three categories of backdoor detection techniques: trigger reversion, model diagnosis and forensic analysis. In trigger reversion, the defender reconstructs the trigger using her own knowledge about the backdoor as constraints to guide trigger searching. For instance, Neural Cleanse~\cite{NeuralCleanse} limits its trigger searching to find those with small norms, and I-BAU~\cite{zeng2021adversarial} approximates the implicit hyper-gradient for trigger reversion. Many proposed backdoor defense techniques~\cite{xiang2020detection, dong2021black, shen2021backdoor, guo2020towards, aiken2021neural} are in this category, based upon different searching algorithms. They are time-consuming, though, when compared with SEAM. 
Model diagnosis looks into the difference between backdoor infected models and benign models. These approaches focus on designing a metric to measure whether a model is closer to a backdoor infected model or a benign model. For example, ABS~\cite{ABS} utilizes a neuron activation vector and MNTD~\cite{xu2021detecting} takes the topological prior and the outputs of a meta-learned model for detecting backdoored models. These approaches are efficient but given the complexity of the neural network, they could be less accurate. 
Forensic analysis is meant to analyze the training data or the operation trace of a model to capture its backdoor behaviors or the attempt to inject a backdoor to the model~\cite{spectral-signatures, strip, SCAn}.  
An example is SCAn~\cite{SCAn}, which leverages the observation that the backdoor-carrying inputs and benign inputs from the source class have significant differences in the representation space. A statistical test can then differentiate them with a theoretical guarantee. A weakness of the approach is requirement for the presence of trigger-carrying inputs. 

Compared with the detection approaches, another line of research is to remove the backdoor from the model, either using recovered backdoor triggers~\cite{tabor} or through blind unlearning~\cite{NAD}. \OURMETHOD is a blind unlearning technique, which we demonstrate to be more effective and efficient than existing approaches, as elaborated in \S~\ref{subsec:comparison}.   

\vspace{-3mm}

\ignore{
There are mainly three categories of backdoor detection techniques: trigger reversion, model diagnosis, and model reconstruction. In trigger reversion, the defender reconstructs the trigger using her own knowledge about the trigger as constraints to guide trigger searching. For instance, Neural Cleanse~\cite{NeuralCleanse} limits its trigger searching to find those with small norms, and I-BAU~\cite{zeng2021adversarial} approximates the implicit hyper-gradient for trigger reversion. Many proposed backdoor defense techniques~\cite{xiang2020detection, dong2021black, shen2021backdoor, guo2020towards, aiken2021neural} are in this category, based upon different searching algorithms. They are time-consuming, though, when compared with SEAM. 
Model diagnosis looks into the difference between backdoor infected models and benign models. These approaches focus on designing a metric to measure whether a model is closer to a backdoor infected model or a benign model. For example, ABS~\cite{ABS} utilizes a neuron activation vector and MNTD~\cite{xu2021detecting} takes the topological prior and the outputs of a meta-learned model for detecting the backdoored model.
Another diagnosis-based defense SCAn~\cite{SCAn} is based upon the observation that the backdoor-carrying inputs and benign inputs from the source class have significant differences in the representation space. A statistical test can differentiate them with a theoretical guarantee. Such diagnosis-based defense is efficient but given the complexity of the neural network, they could be less accurate than trigger reversion. 
Model reconstruction looks into the model parameters mutation. For example Fine-Purning~\cite{FinePruning} is a typical model reconstruction method to unlearn the backdoor, which consists of two steps: 1) pruning the suspicious neurons, and 2) fine-tune the model. 

Compared with these detection, another line of research is to remove the backdoor from the model, either using recovered backdoor triggers~\cite{tabor} or through blind unlearning~\cite{NAD}. \OURMETHOD is a blind unlearning techniques, which compared with prior approaches, is much simpler yet more effective and efficient, as discovered in our study (see our comparison in \S~\ref{subsec:comparison}). 
}

\section{Discussion}\label{Discussion_section}
\vspace{-4mm}


\noindent\textbf{Limitations}. 
Our SEAM is meant to disable a backdoor within an infected model through partially unlearning it, rather than completely remove the backdoor from the model. Particularly, the partial unlearning performed by SEAM suppresses backdoor effects of an infected model in a blind and efficient way, instead of removing all backdoor traces from the model.
As shown on Fig.~\ref{fig:CKA}, our approach breaks the trigger activation chain injected by the adversary into the deep layers (close to the output) of an ML model while largely retaining the model's features in its shallow layers (close to the input). As a result, it may be possible for the adversary to revive the dormant backdoor within the unlearned model, particularly when the model is fine-tuned on the dataset with trigger-carrying inputs. 
To understand the risk, we constructed experiments to investigate how many trigger-carrying inputs need to be injected into the fine-tuning dataset can revive the dormant trigger by our SEAM. We defer the experimental details in Appendix 3. The results (Fig.~\ref{fig:after_recover}) demonstrate that the adversary can revive an effective backdoor only when he can pollute $\ge 4\%$ fine-tuning data, which however is less possible according to our threat model.


Fundamentally, SEAM leverages the architectural property of today's mainstream DNNs, which utilize the same architecture to learn multiple tasks. So, if the adversary manages to separate the backdoor task from the primary task on the architectural level, our protection could fail. For example, one could train a model with two separate DNNs, one for the primary task and the other for the backdoor task~\cite{BadNet}. The model switches between these two networks based upon the trigger pattern recognized from the input. Although this attack can indeed defeat our unlearning, it requires the full control on the training process and therefore cannot be executed through data poisoning. Also a direct combination of two models makes the model architecture differ significantly from the standard ones, rendering the attack easy to detect~\cite{BadNet}. Further research is needed to understand whether other more effective poisoning attacks exist to pose a credible threat to our approach.

\noindent\textbf{Future work}. 
SEAM is meant to be a blind unlearning technique. However, its performance could be improved by leveraging the prior knowledge of a backdoor: e.g., the information about the trigger pattern could help design a more precise forgetting step, which strategically randomizes the labels of a subset of inputs, so as to speed up the step and enhance the ACC the recovery step could achieve. Further, an improvement of SEAM could enable the \textit{forgetting} step to keep track of the speed of degradation for each class, which can be leveraged by the \textit{recovery} step to retrain the model more heavily on selected classes, to make the unlearning process more efficient. Essentially, the \textit{forgetting} step could be viewed as an attempt to find a good initialization point for learning the primary task. This implies that a good initialization may help reduce the ASR of a certain backdoor, which is an open problem for further research. 



\vspace{-4mm}

\section{Conclusion}\label{Conclusion_section}
\vspace{-4mm}
We present SEAM, a novel and high-performance blind unlearning technique for disabling backdoor, and analyzed its effectiveness through experimental studies and theoretic analysis. Our analysis shows that our \textit{forgetting} step actually maximizes the CF on an unknown backdoor in the absence of triggered inputs. Through extensive experiments, we demonstrated efficacy and efficiency of \OURMETHOD on eight datasets with various model architectures against two representative attacks. The results show that \OURMETHOD outperforms existing defenses and achieves a high \textit{Fidelity} efficiently.

\vspace{-10pt}
\section{Acknowledgment}
\vspace{-5pt}

We would like to thank the anonymous reviewers for their insightful comments. This work is partially supported by of IARPA’s TrojAI project (Grant No. W91NF-20-C-0034).

\bibliographystyle{plain}
\bibliography{main}

\appendix
\vspace{-2mm}
\subsection{Theoretical proofs}
\vspace{-4mm}
In this section, we give the proof of the lemmas and theorems for the multi-class classification tasks. The same results also hold for binary classification tasks, for which the proofs are given in an online document \footnote{\url{https://drive.google.com/file/d/1bPT0dbu0cVeBH7qRzAa76nEqyQ27Zh66/view?usp=sharing}} 
(along with the detailed explanation about the independence between the residual and the target task) due to the space limit. Note that in Section~\ref{Theoretical Analyses_section}, the results are illustrated using the binary classification tasks.

\vspace{-1mm}
\section{Theoretical proof}
We consider the multi-class classification problem on $L$ classes. We define $L$ binary classifiers $\{ f^{(j)*}\}^{j \in \{ 1,2,...,L \} }$, in which the $j^{th}$ classifier $f^{(j)*}$ outputs the probability of a given input to be in the $j^{th}$ class. 
For a dataset $X$ with $n$ samples in $L$ classes, $X=\{(x^{(i)}, y^{(i)})\}_{i=1}^{n}$, the label $y^{(i)}$ of the $i^{th}$ input is a vector containing $L$ values, i.e., $y^{(i)}=[y^{(i,1)}, y^{(i,2)}, ..., y^{(i,L)}]$, and $y^{(i,1)}, y^{(i,2)}, ..., y^{(i,L)} \in \{ 0,1\}$. In this multi-class setting, we define the Catastrophe Forgetting (CF) as the following.

\vspace{3pt}\noindent\textbf{Definition of CF in a multi-class classifier}:
The CF from task $\tau_P$ to task $\tau_F$ w.r.t the dataset $X_{\tau_T}$ is:
\begin{equation}
\tiny\Delta^{\tau_{P} \rightarrow \tau_{F}}\left(X_{\tau_{P}}\right) =\sum_{x \in \mathcal{D}_{\tau_{P}}}\left(f_{\tau_{F}}^{(k)\star}(x)-f_{\tau_{P}}^{(k)\star}(x)\right)^{2}
\end{equation}
\noindent where $k \in K = \{ k | f_{\tau_{P}}^{(k)\star}(x) = \max\limits_{1\leq j \leq L} f_{\tau_{P}}^{(j)\star}(x)\}$.   

For the sake of simplicity, we denote the symbol $\ominus$ as the operator:$f_{\tau_{F}}^{\star}(x) \ominus f_{\tau_{P}}^{\star}(x) = f_{\tau_{F}}^{(k)\star}(x)-f^{(k)\star}_{\tau_{P}(x)}$. Thus,
\begin{equation}
\tiny\Delta^{\tau_{P} \rightarrow \tau_{F}}\left(X_{\tau_{P}}\right) = \sum_{x \in \mathcal{D}_{\tau_P}}\left(f_{\tau_{F}}^{\star}(x) \ominus f_{\tau_{P}}^{\star}(x)\right)^{2}
\end{equation}
This definition of CF for multi-class classifiers accounts for the change of confidence from $\tau_{P} \rightarrow \tau_{F}$, w.r.t the desirable class of an input sample for the task $\tau_P$, that is, how much confidence is reduced for the desirable class in $\tau_P$ (equivalent to the confidence increased for the other classes) via the learning procedure $\tau_{P} \rightarrow \tau_{F}$.

Note that the residual is the only term that is different in the definition of the CF for multi-class classification tasks comparing with the definition of the CF for the binary classification tasks (see the online document). The residual term of the multi-class classification tasks can be written as:

$$
\tilde{y}_{\tau_{F}}=y_{\tau_{F}} \ominus f_{\tau_{P}}^{\star}\left(X_{\tau_{F}}\right)
$$

With the CF definition above, we can show that both the upper bound and the lower bound of CF is proportional to  $||\tilde{y}_{\tau_{F}}||^2$ (Lemma 4.2), and when $||\tilde{y}_{\tau_{F}}||^2$ reach maximum, both the upper bound and the lower bound of the CF reach maximum (Theorem 4.3). 


\vspace{3pt}\noindent\textbf{Lemma 4.1.5}
    Let $M \in \mathbb{R}^{n \times n}$ be a symmetric and non-singular matrix, and $v \in \mathbb{R}^n$ is a vector. Then $\lambda_{min}^2 ||v||^2 \leq ||Mv||^2_2 \leq \lambda^2_{max}||v||^2$.

\vspace{3pt}\noindent\textbf{Lemma 4.2 (multi-class version)} 
    For any specific sample $X$, both the upper bound and the lower bound of CF from the source model $f^{\star}{\tau_P}$ to a competitive model trained on $X{\tau_F}$ is proportional to the norm of residual: $||\tilde{y}_{\tau_F}||^2 = {y}_{\tau_F} -f^{\star}_{\tau_P}(X{\tau_F})$.

\begin{proof}    

To show that CF is proportional to residual, we only need to show that for all elements in the residual, $\tilde{y}_{\tau_F}$ is greater than $0$.  
    
\begin{equation}
    \tiny\begin{aligned}
    & \Delta^{\tau_{P} \rightarrow \tau_{F}}\left(X\right) 
    =&\left\| \underbrace{ \phi\left(X\right) \phi\left(X_{\tau_{F}}\right)^{\top}}_{A} \underbrace{\left[\phi\left(X_{\tau_{F}}\right) \phi\left(X_{\tau_{F}}\right)^{\top}+\lambda I\right]^{-1}}_B \tilde{y}_{\tau_{F}}\right\|_{2}^{2}
    \end{aligned}
\end{equation}
    
    \noindent, where $A$ and $B$ are two symmetric and non-singular kernel matrices.
    
    Let $\lambda^2_{(A),min}$ and $\lambda^2_{(A),max}$ be the smallest and largest eigen-value of A, and 
    $\lambda^2_{(B),min}$ and $\lambda^2_{(B),max}$ be the smallest and the largest eigen-value of B. Applying Lemma 4.1.5, we have:
\begin{equation}
\scriptsize\lambda^2_{(A),min} B ||\tilde{y}_{\tau_{F}}||^2 \leq \Delta^{\tau_{P} \rightarrow \tau_{F}} \leq \lambda^2_{(A),max} B ||\tilde{y}_{\tau_{F}}||^2
\end{equation}
    
    \noindent Then applying Lemma 4.1.5 again:
\begin{equation}
\scriptsize\lambda^2_{(A),min} \lambda^2_{(B),min} ||\tilde{y}_{\tau_{F}}||^2 \leq \Delta^{\tau_{P} \rightarrow \tau_{F}}\leq \lambda^2_{(A),max} \lambda^2_{(B),max} ||\tilde{y}_{\tau_{F}}||^2
\end{equation}    

Therefore, both the upper bound and the lower bound of CF is proportional to the residual $||\tilde{y}_{\tau_{F}}||^2$.
    
\end{proof}

\vspace{3pt}\noindent\textbf{Theorem 4.3 (multi-class version)}
Given a fixed input of a training dataset $X_{\tau_F}$, the randomly assigned wrong label ${y}_{\tau_F}$ maximizes the norm of residual $||\tilde{y}_{\tau_F}||^2$, and thus maximizes both the upper bound and the lower bound of CF for any input sample $X$ from the source model to the competitive model trained on the labeled dataset $D_{\tau_F} = (X_{\tau_F}, y_{\tau_F})$.

\begin{proof}    
    Here, we prove how the randomly assigned wrong label $y_{\tau_F}$ maximizes the upper bound of CF. The proof for the lower bound is similar, and thus is not shown. 
    
    Lemma 4.2 shows that the upper bound of CF is proportional to $||\tilde{y}_{\tau_{F}}||^2$, i.e., $ \sup{\Delta^{\tau_{P} \rightarrow \tau_{F}}} \propto ||\tilde{y}_{\tau_{F}}||^2$. Thus the maximum of $||\tilde{y}_{\tau_{F}}||^2$ leads to the maximum of $\sup{\Delta^{\tau_{P} \rightarrow \tau_{F}}}$. 
    
    Next, we show the randomly assigned wrong labels $y_{\tau_F}$ can maximize $||\tilde{y}_{\tau_{F}}||^2$.
    Recall the residual term in the CF of multi-class classification models is defined as:
    $\tilde{y}_{\tau_{F}}=y_{\tau_{F}} \ominus f_{\tau_{P}}^{\star}\left(X_{\tau_{F}}\right)$.
   Let $(x^{(i)}_{\tau_F}, y^{(i)}_{\tau_F})$ be the $i^{th}$ data point in the training dataset of the task $\tau_F$. $|| \tilde{y}^{(i)}_{\tau_F}||^2 = ||y^{(i,k)}_{\tau_{F}} - f_{\tau_{P}}^{(k)\star}\left(x^{(i)}_{\tau_F}\right)||^2$ is the residual norm of $x^{(i)}_{\tau_F}$ from task $\tau_P$ to task $\tau_F$. 
   Specifically, the first term in the residual norm is $y^{(i,k)}_{\tau_{F}}$, the $k^{th}$ label of $x_{\tau_F}^{(i)}$, and $y^{(i,k)}_{\tau_{F}} \in \{ 0,1 \}$. 
   The second term is $f_{\tau_{P}}^{(k)\star}\left(x^{(i)}_{\tau_F}\right)$, the predicted outcome for the $k^{th}$ class by the model for the task $\tau_P$ on $x_{\tau_{F}}^{(i)}$, and $f_{\tau_{P}}^{(k)\star}(x^{(i)}_{\tau_F}) \in [ 0,1 ]$. 
   Since in the task $\tau_F$, the label of the $i^{th}$ input is the randomly wrong label, i.e., $y_{\tau_F}^{(i,k)} = 0 \neq 1 = y_{\tau_P}^{(i,k)}$, and the $i^{th}$ input is the same as the $i^{th}$ input of the task $\tau_P$, i.e., $x_{\tau_F}^{(i)} = x_{\tau_P}^{(i)}$, we have $f_{\tau_{P}}^{(k)\star}(x^{(i)}_{\tau_F}) = f_{\tau_{P}}^{(k)\star}(x^{(i)}_{\tau_P}) = y_{\tau_P}^{(i,k)} = 1$ for a well trained classifier $f_{\tau_P}^{(k)\star}$. Thus,  $ ||y^{(i,k)}_{\tau_{F}} - f_{\tau_{P}}^{(k)\star}\left(x^{(i)}_{\tau_F}\right)||^2 = 1$ reaches the maximum, i.e., the theorem is proved.
\end{proof}

\vspace{-3mm}
\subsection{Dataset}
\label{subsec:datasets}
\vspace{-4mm}
\begin{table*}[htp]
\centering
\caption{Datasets statistics. The training samples and testing samples of models in \textit{Round 3-7} of TrojaAI competition are different one by one, and the details are stored in configuration file associated with each model. Here, we only list the approximate value for reference.}
\label{tab:datasets}
\begin{adjustbox}{width=0.65\textwidth}
\begin{tabular}{|l|c|c|c|c|c|c|c|c|}
\hline
                      & \multicolumn{1}{l|}{MNIST} & \multicolumn{1}{l|}{GTSRB} & \multicolumn{1}{l|}{CIFAR10} & \multicolumn{1}{l|}{Round 3} & \multicolumn{1}{l|}{Round 4} & \multicolumn{1}{l|}{Round 5} & \multicolumn{1}{l|}{Round 6} & \multicolumn{1}{l|}{Round 7} \\ \hline
Models (\#)           & 12                         & 12                         & 12                           & 1584                         & 1584                         & 2664                         & 3114                         & 960                          \\ \hline
Training samples (\#) & 60,000                     & 39,209                     & 50,000                       & $\sim$40,000                & $\sim$40,000                & $\sim$100,000                & $\sim$40,000                & $\sim$40,000                \\ \hline
Testing samples (\#)  & 10,000                     & 12,630                     & 10,000                       & $\sim$4,000                 & $\sim$4,000                 & $\sim$10,000                 & $\sim$4,000                 & $\sim$4,000                 \\ \hline
\end{tabular}
\end{adjustbox}
\end{table*}

\vspace{3pt}\noindent\textbf{TrojAI competition datasets}. 
TrojAI is a competition founded by IARPA for backdoor detection\ignore{ on different domains under various circumstances}. This competition includes several rounds that are designed for different image processing and NLP tasks. In each round, a set of training datasets, testing datasets and holdout datasets are published, which contain benign and infected models with various structures. In our research, we evaluated \OURMETHOD on all three datasets released for \textit{Round 3-7}. Specifically, \textit{Round 3-4} are designed for backdoor detection on image classification tasks and \textit{Round 5-7} for backdoor detection on NLP tasks. 

The dataset of \textit{Round 3} has totally 1584 models with 22 different model structures in 7 types of architectures: ResNet $\{18$, $34$, $50$, $101$, $152\}$~\cite{resnet}, wide-ResNet $\{50$, $101\}$~\cite{wide-resnet}, DenseNet $\{121$, $161$, $169$, $201\}$~\cite{densenet}, Inception $\{v1$, $v3\}$~\cite{inception}, SqueezeNet $\{v1.0$, $v1.1$, $v2\}$~\cite{squeezenet}, ShuffleNet $\{1.0$, $1.5$, $2.0\}$~\cite{shufflenet}, and Vgg $\{11bn$, $13bn$, $16bn\}$~\cite{vggnet}. Backdoors injected into these models can be universal (mapping all labels to a target one) or specific (mapping a specific source label to a target), and multiple backdoors can present in one model. The triggers of the backdoors include pixel patterns (e.g., polygons with solid color) and Instagram filters that are applied to images to change their styles (e.g., Gotham Filter).

The dataset of \textit{Round 4} contains 1584 models with 16 structures. Triggers in this round are more subtle than those in \textit{Round 3}. They could be spatially dependent, only taking effect when they appear at specific locations relative to the foreground objects in images, and spectral dependent, requiring a right combination of colors in order to cause misclassification.

The dataset of \textit{Round 5} includes 2664 sentiment classification models trained on the IMDB movie review dataset~\cite{maas-EtAl:2011:ACL-HLT2011} and the Amazon review dataset~\cite{ni2019justifying} in 3 popular NLP model architectures: BERT~\cite{bert}, GPT-2~\cite{gpt-2} and DistilBERT~\cite{distilbert}. A trigger in this round can be a character (e.g., ``@''), a word (e.g., ``cromulent'') or a phrase (e.g., ``I watched an 3D movie'').  The dataset of \textit{Round 6} share the same settings except that its dataset for training detector is very small with only 48 models. Thus, together with the training, testing and holdout datasets, we used 3114 models of \textit{Round 6} in our experiments. 


The dataset of \textit{Round 7} carries 960 models for Named-Entity Recognition (NER). These models trained on BBN~\cite{weischedel2005bbn}, CoNLL-2003~\cite{10.3115/1119176.1119195} and OntoNotes~\cite{hovy-etal-2006-ontonotes} NER datsets, and are in 4 different model architectures: BERT~\cite{bert}, DistilBERT~\cite{distilbert}, RoBERTa~\cite{roberta} and MobileBERT~\cite{mobilebert}. Different from binary classification in \textit{Round 6}, the NER task of \textit{Round 7} involves multiple classes, which assign each word into one of the several categories. It uses the trigger settings of \textit{Round 6}.

\vspace{3pt}\noindent\textbf{MNIST, GTSRB, CIFAR10}. These three datasets are constructed for image classification tasks: MNIST is for recognizing handwritten digits, GTSRB is for detecting traffic signs and CIFAR10 is for classifying general images. On these datasets,  we trained 3 models for each of the 4 model architectures that includes ShuffleNetx1.0, Vgg16, ResNet18 and ResNet101. Totally 12 models trained on each dataset were used for backdoor injections (e.g., TrojanNet~\cite{trojannet}) to get backdoor infected models.

\subsection{Additional Experiments}
\label{app:add_exp}

\vspace{3pt}\noindent\textbf{Backdoor-revival experiments}.

\begin{figure}[htbp]
\centering
\includegraphics[width=6cm]{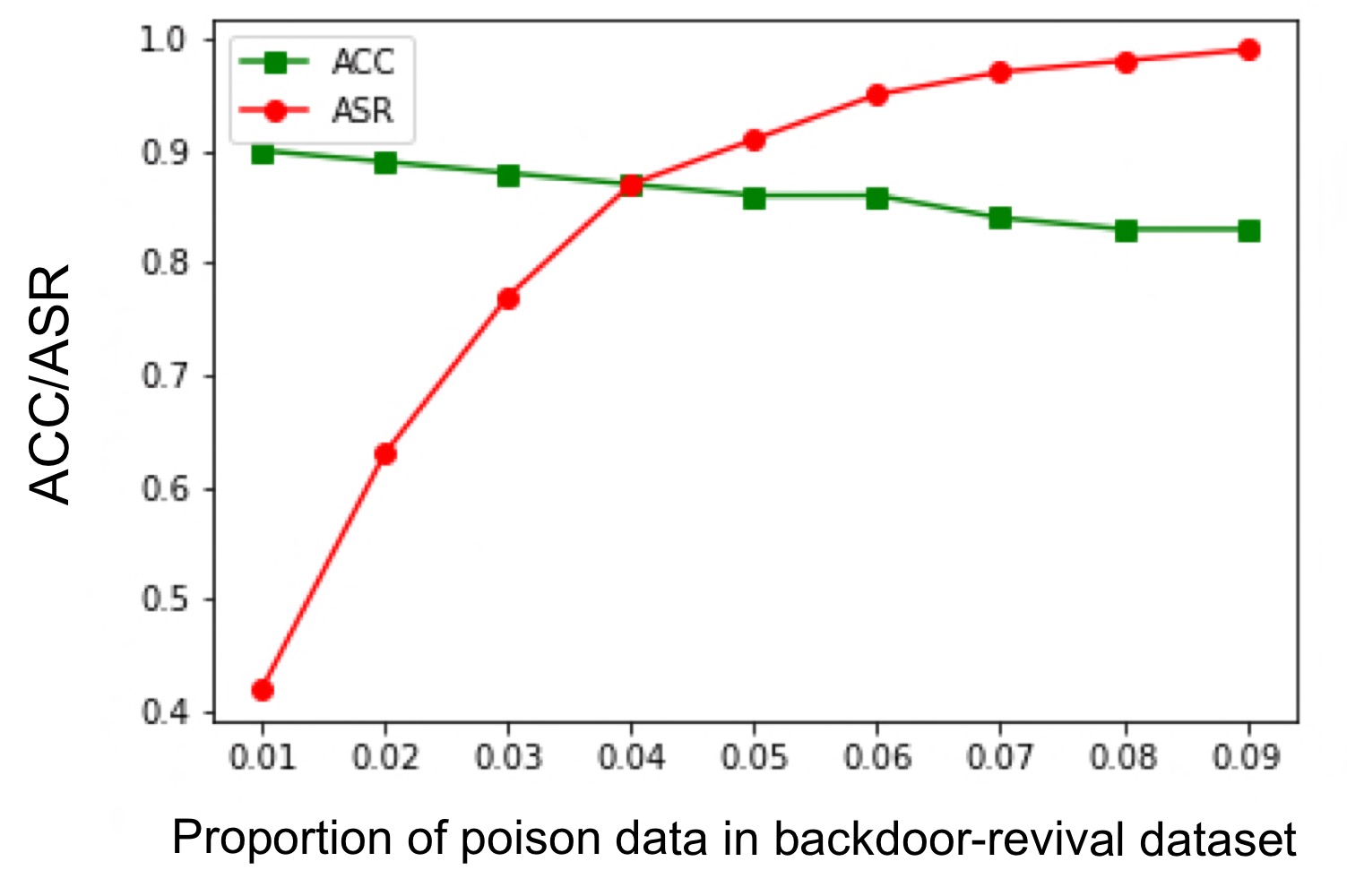}
\caption{ASR and ACC changes when the adversary fine-tunes an unlearned model on a backdoor-revival dataset with different poison proportions.}
\label{fig:after_recover}
\end{figure}

We ran SEAM on the backdoored models in four mainstream architectures (ShuffleNetx1.0, Vgg16, ResNet18, and ResNet101), and further fine-tuned the unlearned models on a poisoned dataset (referred to as the ``backdoor-revival'' dataset). Specifically, we constructed the backdoor-revival dataset by randomly sampling 10,000 benign images from CIFAR10 used to train these models and replacing a portion (1\% - 9\%) of them with trigger-carrying instances. Our experimental results are presented in Fig.~\ref{fig:after_recover}. As we can see from the figure, the backdoors within these models can be revived (ASR $\ge 89\%$) through fine-tuning them on the backdoor-revival dataset when the portion of trigger-carrying inputs grows above $4\%$.

\begin{figure}[h]
\vspace{-1.0em}
\centering
\includegraphics[width=0.3\textwidth]{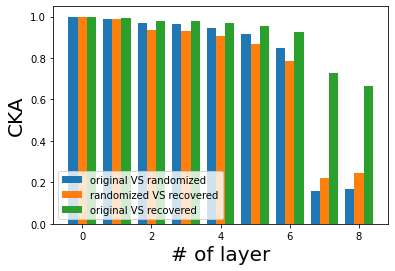}
\caption{CKA on each layer of ResNet-18. Layer 0 is the first layer taking the image as its input and layer 9 is the last second layer.}
\label{fig:CKA_2}
\vspace{-1em}
\end{figure}

\vspace{3pt}\noindent\textbf{CKA experiments of EW on CIFAR100}.
On CIFAR100, we ran SEAM on a EW infected ResNet-18 model with a clean dataset with the size of $0.1\%$ of the training dataset for forgetting and a clean dataset with the size of $10\%$ of the training dataset for recovery, and calculated the CKA for each layer. The results are demonstrated on Fig.~\ref{fig:CKA_2}. We observe that, even the backdoor has been entangled with the primary task by the EW, there is still chance to depress this backdoor without harming the primary task performance of the target model through SEAM that preserves many features of the original model on shallow layers. 

\end{document}